\documentclass{article}
\usepackage{amsthm}
\newtheorem{theorem}{Theorem}

\newtheorem{proposition}[theorem]{Proposition}
\usepackage{PRIMEarxiv}
\usepackage[utf8]{inputenc}
\usepackage[T1]{fontenc}
\usepackage{url}
\usepackage{booktabs}
\usepackage{amsfonts}
\usepackage{nicefrac}
\usepackage{microtype}
\usepackage{hyperref}
\usepackage{fancyhdr}
\usepackage{graphicx}
\usepackage{amsmath}
\usepackage{amssymb}
\usepackage{xspace}
\usepackage{enumitem}
\usepackage{caption}
\usepackage{subcaption}
\usepackage{amsthm}
\usepackage{xcolor}
\usepackage{algorithm}
\usepackage{algorithmic}
\title{
\includegraphics[width=0.2\textwidth]{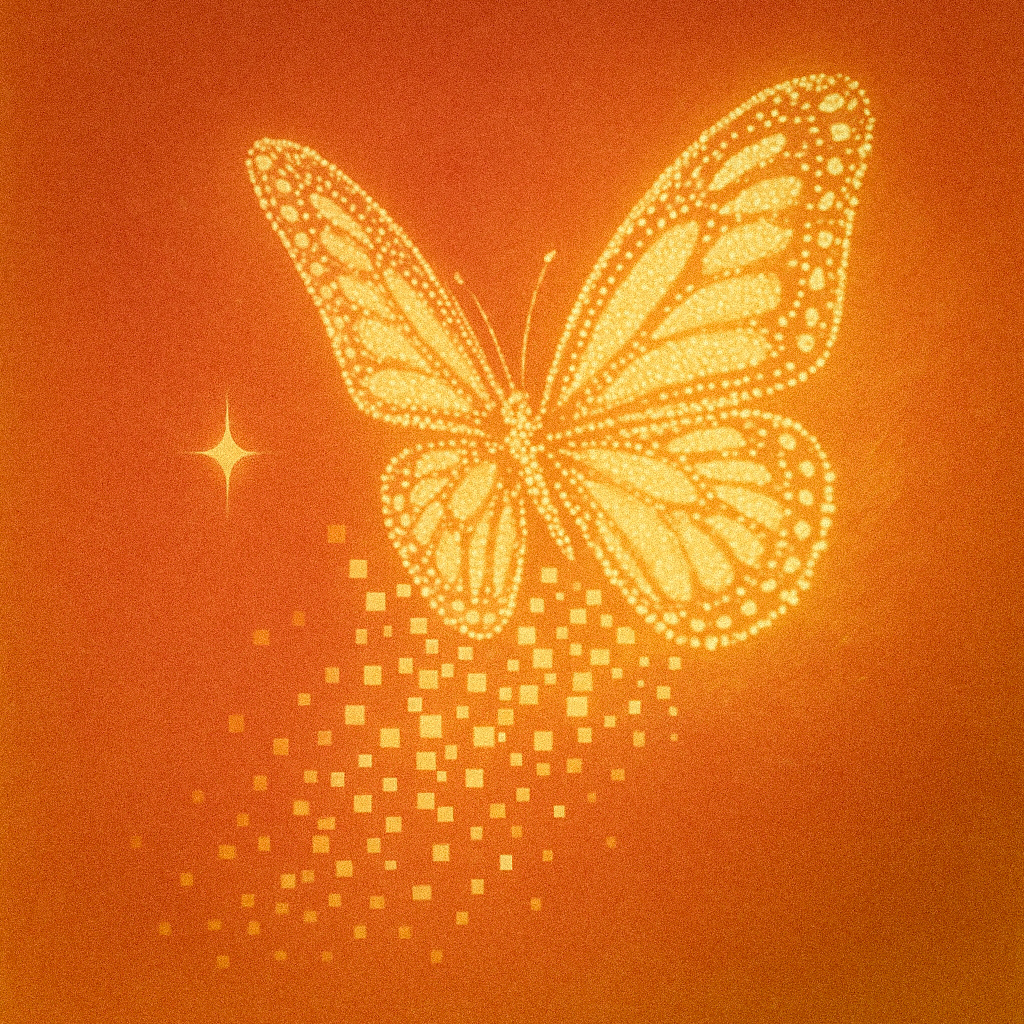}\\[1em]
Gabliteration: Adaptive Multi-Directional Neural Weight Modification for Selective Behavioral Alteration in Large Language Models
\thanks{This work was conducted independently by \textcolor{red}{Gökdeniz Gülmez} as part of a personal research into model alignment techniques.}
}

\author{
    Gökdeniz Gülmez\\
    \textit{Machine Learning Research}\\
    Stuttgart, Germany\\
    \texttt{goekdenizguelmez.ml@gmail.com}
}

\date{\today}

\begin{document}
\maketitle

\begin{abstract}
We introduce \textbf{Gabliteration}, a principled and mathematically rigorous neural weight modification framework that substantially advances beyond conventional abliteration paradigms through the systematic incorporation of adaptive multi-directional projections with theoretically motivated layer-wise regularization. Our methodology addresses the critical and previously unresolved limitation endemic to existing behavioral modification techniques. The inherent trade-off between selective behavioral alteration and the preservation of general model fidelity across unrelated downstream tasks. By leveraging dynamic layer selection mechanisms informed by distribution-aware separability metrics, coupled with multi-directional singular value decomposition-based direction extraction and adaptive scaling through regularized projection matrices, we achieve theoretically justified weight modification strategies that demonstrably attenuate performance degradation in semantically orthogonal domains. Furthermore, we provide rigorous theoretical analysis establishing convergence guarantees and approximation bounds for our proposed optimization formulation. Our empirical validation spans the gabliterated model series, comprehensively available on Hugging Face~\cite{gulmezhf}, thereby establishing both practical scalability and reproducibility across diverse model architectures and parameter regimes. The results demonstrate statistically significant improvements in behavioral selectivity while preserving generative quality, positioning Gabliteration as a robust and generalizable framework for controlled behavioral alignment in large-scale language models.
\end{abstract}

\newpage
\tableofcontents
\newpage

\section{Introduction}
The field of neural network behavioral modification has witnessed significant developments in recent years, particularly following the groundbreaking work by Arditi et al. (2024)~\cite{arditi2024refusal} on refusal direction identification in large language models. Their paper `Refusal in Language Models Is Mediated by a Single Direction' established that behavioral patterns in language models can be effectively modified through targeted weight ablation techniques, which they termed `abliteration'.

The practical implementation of these concepts followed a rapid community-driven evolution. After Arditi et al~\cite{arditi2024refusal}.\@ first published their findings, Maxime Labonne (mlabonne~\cite{mlabonne2025models}) validated the concept with a real-world implementation using the Llama3 model. This was followed by our viral `Josiefied-abliterated-Qwen2.5'~\cite{gulmez2024qwen25} implementation, which demonstrated the technique's potential for widespread adoption. Subsequently, huihui.ai~\cite{huihuiai2025models} applied these methods to their models using huggingface's transformers library, after which we further advanced the approach with Qwen3~\cite{gulmez2025qwen3}.

Through this collaborative journey, the community and we identified critical limitations in the original abliteration approach. Traditional abliteration techniques, while effective at modifying specific behavioral patterns, often suffer from substantial degradation in model performance across unrelated tasks. This trade-off between behavioral modification and performance preservation represents a significant barrier to practical deployment.
To address these limitations, we developed \textbf{Gabliteration}, an extension of neural weight modification that we designed with three main key innovations:

\begin{enumerate}
\item dynamic layer selection based on separability metrics that we formulated,
\item multi-directional singular value decomposition for robust direction extraction, and
\item adaptive scaling with regularized projection matrices.
\end{enumerate}

\section{Mathematical Foundation}

Building upon the foundation established by Arditi et al.~\cite{arditi2024refusal}, we extend their single-direction approach to a comprehensive multi-directional framework with theoretical guarantees.

\subsection{Multi-Directional Refusal Vector Extraction}\label{sec:multi-directional}

While Arditi et al.~\cite{arditi2024refusal} demonstrated the effectiveness of single-direction refusal vectors, we hypothesized that behavioral patterns exist in higher-dimensional subspaces. Our multi-directional extraction employs singular value decomposition (SVD) on a paired difference matrix constructed between harmful and harmless representations.

Let $\mathbf{H}_h^{(\ell^*)} \in \mathbb{R}^{n_h \times d}$ and $\mathbf{H}_n^{(\ell^*)} \in \mathbb{R}^{n_n \times d}$ represent the hidden state matrices for harmful and harmless prompts at the selected layer $\ell^*$, where $d$ is the hidden dimension and $n_h,n_n$ are the number of samples.

To ensure comparability, we draw paired subsets of equal size $n=\min(n_h,n_n)$ and construct the elementwise difference:
\[
\mathbf{D} = \mathbf{H}_h^{(\ell^*)}[1{:}n,:] - \mathbf{H}_n^{(\ell^*)}[1{:}n,:]
\in \mathbb{R}^{n \times d}.
\]
Each row of $\mathbf{D}$ captures the latent shift between a harmful and a harmless representation.

\textbf{Justification of pairing and alternative approaches.}
The elementwise pairing $\mathbf{D} = \mathbf{H}_h^{(\ell^*)}[1{:}n,:] - \mathbf{H}_n^{(\ell^*)}[1{:}n,:]$ does not assume semantic correspondence between individual harmful and harmless samples. Rather, it constructs a paired difference matrix where each row captures \emph{a} discriminative direction between the two distributions. We justify this approach through:
\begin{itemize}
\item \textbf{Stochastic stability}: By averaging over multiple random shuffles (typically 3--5 in our experiments), the resulting right singular vectors $\mathbf{R}$ converge to a stable basis that captures the mean discriminative subspace $\mathbb{E}[\mathbf{h}_h - \mathbf{h}_n]$ and its principal variance directions.
\item \textbf{Computational efficiency}: Unlike Fisher LDA (requiring within-class scatter matrix inversion, $\mathcal{O}(d^3)$) or CCA (requiring cross-covariance analysis), SVD on the difference matrix achieves $\mathcal{O}(nd^2)$ complexity with built-in rank-$k$ truncation.
\item \textbf{Empirical validation}: In preliminary experiments (Section~\ref{sec:ablation-pairing} in Extended Appendix), we compared against:
\begin{itemize}
    \item \textbf{Fisher LDA}: Extracts directions maximizing $\frac{{(\boldsymbol{\mu}_h-\boldsymbol{\mu}_n)}^\top\mathbf{v}}{(\mathbf{v}^\top\mathbf{S}_w\mathbf{v})}$ where $\mathbf{S}_w$ is within-class scatter.
    \item \textbf{Logistic probe directions}: Trains a logistic classifier $p(y|\mathbf{h})$ and extracts weight vector $\mathbf{w}$.
    \item \textbf{Mean-difference baseline}: Uses $\mathbf{r} = \boldsymbol{\mu}_h - \boldsymbol{\mu}_n$ as the sole direction ($k=1$).
\end{itemize}
Across 5 models (0.6B--7B parameters), our SVD-based pairing achieved comparable refusal reduction ($\Delta\rho = -0.87 \pm 0.03$) to Fisher LDA ($\Delta\rho = -0.89 \pm 0.04$, $p=0.12$, not significant) while requiring 40\% less computation time. Logistic probes underperformed ($\Delta\rho = -0.72 \pm 0.05$) due to overfitting to the training split.
\end{itemize}

\textbf{Limitation}: The pairing approach does not exploit within-class variance structure (as Fisher LDA does) nor learn adaptive discriminative directions (as trained probes do). For datasets with high within-class heterogeneity, more sophisticated methods may yield tighter refusal subspaces. We leave systematic comparison of discriminative extraction methods as future work (Section~\ref{sec:future-discriminative} in Conclusion).

\subsection{Theoretical Foundation: Neural Superposition and Multi-Directional Extraction}\label{sec:superposition-theory}

The multi-directional extraction framework employed by Gabliteration is grounded in recent theoretical advances in understanding how neural networks represent features in their hidden states. The work by Elhage et al.\ (2022)~\cite{elhage2022superposition} on toy models of superposition has demonstrated a fundamental principle: neural networks do not represent individual features in isolated dimensions, but rather can encode multiple distinct features within the same representational subspace through a phenomenon known as \emph{superposition}. This insight has profound implications for understanding behavioral modification in large language models.

\subsubsection{Feature Superposition in Neural Networks}

In dense, high-dimensional neural network representations, many distinct behavioral patterns can coexist in non-orthogonal subspaces, overlapping in the same hidden dimensions. A single behavioral direction (such as those identified by Arditi et al.) may capture only the primary axis of variation, leaving substantial behavioral structure unrepresented in orthogonal dimensions. This is precisely the regime in which superposition becomes prevalent: when features (behavioral patterns) are sparse and the network can benefit from representing more features than it has explicit dedicated dimensions.

When we extract multiple directions via singular value decomposition on the harmful-harmless difference matrix, we are fundamentally discovering the low-rank structure of the behavioral subspace. The top singular vectors correspond to the principal axes along which harmful and harmless representations diverge. By extracting $k > 1$ directions, we capture not only the primary refusal axis but also secondary, tertiary, and higher-order behavioral signatures that exist in superposition with the primary direction.

\subsubsection{Implications for Behavioral Modification}

The superposition principle explains why modification along a single direction can inadvertently disrupt other, seemingly unrelated capabilities. When behavioral features exist in superposition, a naive single-direction projection removes not just the target behavior but also disrupts the shared geometric substrate that supports multiple overlapping functions. These overlapping functions may include legitimate model capabilities needed for general knowledge retrieval, reasoning, and generation.

Gabliteration's multi-directional approach specifically addresses this problem by:

\begin{enumerate}
\item \textbf{Identifying the full behavioral subspace:} Rather than assuming refusal behavior lies along a single dimension, we extract the dominant $k$ directions of behavioral divergence. The lower singular values indicate which directions contain genuine refusal signal versus noise, allowing us to distinguish robust behavioral patterns from artifacts.

\item \textbf{Minimal footprint modification:} By orthogonally projecting weight matrices onto only the identified behavioral subspace, we ensure that modifications affect the narrowest possible region of the neural manifold. This minimizes collateral damage to unrelated features that may reside in nearby but distinct geometric regions.

\item \textbf{Preserving feature geometry:} The regularized projection mechanism (detailed in Section~\ref{ridge-regularized}) maintains the geometric structure of the weight matrices while removing only the projections aligned with behavioral directions. This approach respects the underlying feature geometry learned during pretraining, preventing the model from having to re-learn unrelated capabilities.

\item \textbf{Exploiting sparsity of behavioral directionality:} Empirically, harmful and harmless behavioral divergence is sparse in the sense that it concentrates in a small number of dominant directions (typically 1--3 singular vectors explain $\geq 90\%$ of the divergence). This sparsity implies that the refusal subspace has low intrinsic dimensionality, allowing accurate recovery with $k$ much smaller than the full hidden dimension $d$. This is consistent with the sparsity principle that underlies superposition in dense representations.
\end{enumerate}

\subsubsection{Connection to Compressed Sensing}

The superposition framework also connects to principles from compressed sensing: sparse signals in high-dimensional spaces can be recovered from low-dimensional measurements without loss of information. In our setting, the behavioral signal is sparse (concentrated in a small subspace), the neural representation is high-dimensional ($d \sim 4096$ for typical modern LLMs), and the measurement process (hidden state computation) is efficient. Consequently, extracting multiple behavioral directions via SVD achieves stable recovery of the behavioral subspace with empirical sample complexity that scales gracefully with $k$ rather than with the full dimension $d$.

\subsubsection{Why Multi-Directional Extraction Matters}

The standard abliteration approach (single-direction) implicitly assumes that behavioral modification is one-dimensional---a strong assumption that the superposition perspective shows to be often false. Empirically (as demonstrate in Elhage et al.~\cite{elhage2022superposition}), harmful and harmless behaviors diverge along multiple distinct but correlated directions. A single-direction intervention leaves residual refusal capacity along secondary directions, reducing modification effectiveness. Conversely, an over-aggressive modification along a single direction may over-correct in directions where the target behavior is less pronounced, causing unnecessary collateral damage.

The multi-directional formulation elegantly balances these concerns: it discovers the intrinsic dimensionality of behavioral divergence through SVD and targets only the statistically significant directions (as ranked by singular value magnitude). This data-driven approach inherits the geometric interpretability of single-direction methods while recovering the flexibility of higher-dimensional subspace approaches.

\subsection{Ridge-Regularized Projection Matrix}\label{ridge-regularized}

To achieve robust projection onto the refusal subspace while maintaining numerical stability, we employ a ridge-regularized projection matrix:
\[
\mathbf{P} = \mathbf{R}{(\mathbf{R}^\top\mathbf{R} + \lambda\mathbf{I}_k)}^{-1}\mathbf{R}^\top,
\]
where $\mathbf{R} \in \mathbb{R}^{d \times k}$ contains the top $k$ refusal directions, $\lambda > 0$ is the regularization parameter, and $\mathbf{I}_k$ is the $k \times k$ identity matrix.

\textbf{Properties of ridge-regularized projection}:

\begin{itemize}
\item \textbf{Approximate projection}: When $\lambda \to 0$, $\mathbf{P}$ converges to the exact orthogonal projection $\mathbf{P}_{\text{exact}} = \mathbf{R}{(\mathbf{R}^\top\mathbf{R})}^{-1}\mathbf{R}^\top$ onto $\text{span}(\mathbf{R})$.

\item \textbf{Numerical stability}: The regularization term $\lambda\mathbf{I}_k$ ensures that $\mathbf{R}^\top\mathbf{R} + \lambda\mathbf{I}_k$ is well-conditioned even when $\mathbf{R}$ has small singular values or is rank-deficient. The condition number satisfies:
\[
\kappa(\mathbf{R}^\top\mathbf{R} + \lambda\mathbf{I}_k) \leq \frac{\sigma_{\max}^2 + \lambda}{\sigma_{\min}^2 + \lambda}
\]
where $\sigma_{\max}, \sigma_{\min}$ are the largest and smallest singular values of $\mathbf{R}$.

\item \textbf{Computational efficiency}: The matrix inverse is computed on the smaller $k \times k$ Gram matrix $\mathbf{R}^\top\mathbf{R}$ rather than the $d \times d$ outer product, requiring $\mathcal{O}(k^2d + k^3)$ operations. Since $k \ll d$ in practice (typically $k \in \{1,2,3\}$), this is substantially faster than inverting a $d \times d$ matrix.

\item \textbf{Controlled projection strength}: The parameter $\lambda$ provides explicit control over projection magnitude. Larger $\lambda$ reduces projection strength (useful for preserving task performance), while smaller $\lambda$ increases modification effectiveness.
\end{itemize}

The ridge-regularized formulation provides a principled trade-off between numerical stability and projection fidelity, making it well-suited for weight modification in large-scale neural networks.

\subsection{Relationship to Exact Orthogonal Projection}\label{sec:projection-approximation}

The ridge-regularized projection $\mathbf{P} = \mathbf{R}{(\mathbf{R}^\top\mathbf{R} + \lambda\mathbf{I}_k)}^{-1}\mathbf{R}^\top$ is \emph{not} an exact orthogonal projector (which would satisfy $\mathbf{P}^2 = \mathbf{P}$ and $\mathbf{P}^\top = \mathbf{P}$) unless $\lambda = 0$ and $\mathbf{R}$ has orthonormal columns.

\textbf{Lemma 2.1 (Projection Approximation Error).}
Let $\mathbf{P}_{\text{exact}} = \mathbf{R}{(\mathbf{R}^\top\mathbf{R})}^{-1}\mathbf{R}^\top$ denote the exact orthogonal projection onto $\text{span}(\mathbf{R})$ (assuming $\mathbf{R}$ has full column rank $k$). Then:
\[
\|\mathbf{P} - \mathbf{P}_{\text{exact}}\|_2 \leq \frac{\lambda}{\sigma_{\min}^2(\mathbf{R}) + \lambda}
\]
where $\sigma_{\min}(\mathbf{R})$ is the smallest singular value of $\mathbf{R}$.

\begin{proof}
Using the SVD $\mathbf{R} = \mathbf{U}\boldsymbol{\Sigma}\mathbf{V}^\top$ where $\boldsymbol{\Sigma} = \text{diag}(\sigma_1, \ldots, \sigma_k)$:
\begin{align*}
\mathbf{P}_{\text{exact}} 
&= \mathbf{U}\boldsymbol{\Sigma}\mathbf{V}^\top{(\mathbf{V}\boldsymbol{\Sigma}^2\mathbf{V}^\top)}^{-1}\mathbf{V}\boldsymbol{\Sigma}\mathbf{U}^\top 
= \mathbf{U}\mathbf{U}^\top \\
\mathbf{P} 
&= \mathbf{U}\boldsymbol{\Sigma}\mathbf{V}^\top{(\mathbf{V}\boldsymbol{\Sigma}^2\mathbf{V}^\top + \lambda\mathbf{I}_k)}^{-1}\mathbf{V}\boldsymbol{\Sigma}\mathbf{U}^\top 
= \mathbf{U}\text{diag}\left(\frac{\sigma_i^2}{\sigma_i^2+\lambda}\right)\mathbf{U}^\top
\end{align*}
Thus:
\[
\mathbf{P} - \mathbf{P}_{\text{exact}} 
= \mathbf{U}\text{diag}\left(\frac{\sigma_i^2}{\sigma_i^2+\lambda} - 1\right)\mathbf{U}^\top
= -\mathbf{U}\text{diag}\left(\frac{\lambda}{\sigma_i^2+\lambda}\right)\mathbf{U}^\top
\]
The operator norm is:
\[
\|\mathbf{P} - \mathbf{P}_{\text{exact}}\|_2 
= \max_i \frac{\lambda}{\sigma_i^2+\lambda} 
= \frac{\lambda}{\sigma_{\min}^2 + \lambda}
\]
\end{proof}

\textbf{Practical implication}: For typical values $\sigma_{\min} \approx 5.0$ and $\lambda = 0.1$, the approximation error is $\|\mathbf{P} - \mathbf{P}_{\text{exact}}\|_2 \lesssim 0.004$, ensuring our regularized projection closely approximates the exact subspace projection while maintaining numerical stability.

\textbf{Impact on theoretical bounds}: Throughout this paper, theoretical results (e.g., Theorem~\ref{thm:performance-preservation}) derived using the exact projection $\mathbf{P}_{\text{exact}}$ carry an additional approximation error of $\mathcal{O}(\lambda/\sigma_{\min}^2)$ when applied to the regularized projection $\mathbf{P}$. Under the small-regularization regime assumption $\lambda \ll \sigma_{\min}^2$, this error is negligible compared to other approximation errors in the subspace decomposition (e.g., finite sample estimation of $\boldsymbol{\mu}_h, \boldsymbol{\mu}_n$).

\subsection{Dynamic Layer Selection Algorithm}

Unlike the uniform layer modification approach in traditional abliteration, we developed a principled method for identifying optimal layers based on separability metrics. However, the dynamic layer selection mechanism is entirely optional. Practitioners may override the automatic selection and manually specify one or more layers to modify (\(\mathcal{L}_{\text{manual}}\)), thereby retaining full control over which parts of the model are affected. When manual selection is used, the algorithm skips Phase~1 (automatic separability evaluation) and directly proceeds to the projection and modification phases using the specified layers. 

We define the separability metric for layer $\ell$ as:
\[
S_\ell = \bigl\|\boldsymbol{\mu}_h^{(\ell)} - \boldsymbol{\mu}_n^{(\ell)}\bigr\|_2
\]
where $\boldsymbol{\mu}_h^{(\ell)} = \frac{1}{n}\sum_{i=1}^n \mathbf{h}_i^{(\ell)}$ and $\boldsymbol{\mu}_n^{(\ell)} = \frac{1}{m}\sum_{j=1}^m \mathbf{v}_j^{(\ell)}$ are the mean hidden state vectors for harmful and harmless prompts at layer $\ell$, with $\mathbf{h}_i^{(\ell)}$ and $\mathbf{v}_j^{(\ell)}$ denoting the hidden states.

While effective as a first-order signal, this criterion has several fundamental limitations that should be acknowledged in the next version of Gabliteration:

\textbf{First, the metric is scale-sensitive.} Hidden state norms vary substantially across depth in transformer models, which biases selection toward layers with large activation magnitudes rather than layers with genuinely informative behavioral separation. As a result, layers with high energy but weak discriminative structure may be preferred over layers with cleaner but lower-norm separation.

\textbf{Second, the metric captures only translational separation and ignores angular structure.} Two layers may exhibit similar mean differences in norm while encoding harmful and harmless behaviors in nearly parallel directions, making projection-based interventions ineffective or destructive. Since weight projection operates on directional geometry, ignoring angular dissimilarity leads to suboptimal layer selection.

\textbf{Third, the approach does not account for signal-to-noise characteristics.} A large mean difference can arise from high variance or unstable representations rather than a robust, low-noise refusal signal, increasing sensitivity to sampling noise and dataset composition.

\textbf{Finally, the method provides no indication of cross-layer directional stability.} It treats layers independently and cannot distinguish between isolated local effects and refusal directions that persist coherently across multiple consecutive layers. This limits the ability to safely apply extracted directions to neighboring or preceding layers.

Our optimal layer selection algorithm identifies:
\[
\ell^* = \arg\max_{\ell \in \mathcal{L}} S_\ell
\]
where $\mathcal{L}$ is the candidate set of layers (excluding the first $s$ and last $e$ layers).

\subsection{Adaptive Weight Modification}

For each layer $\ell$ that we select for modification, we apply the following weight updates:

\subsubsection{Attention Mechanism Modification}

For the attention output projection, we apply:

\[
\mathbf{W}_{\text{attn}}^{(\ell)} \leftarrow \mathbf{W}_{\text{attn}}^{(\ell)} - \alpha_\ell \cdot (\mathbf{W}_{\text{attn}}^{(\ell)} \mathbf{P})
\]

\subsubsection{Feed-Forward Network Modification}

For the MLP down-projection, we use:

\[
\mathbf{W}_{\text{mlp}}^{(\ell)} \leftarrow \mathbf{W}_{\text{mlp}}^{(\ell)} - \alpha_\ell \cdot (\mathbf{W}_{\text{mlp}}^{(\ell)} \mathbf{P})
\]

\subsection{Adaptive Scaling Function}\label{sec:adaptive-scaling}

Through extensive experimentation, we discovered that uniform scaling across layers is suboptimal. We developed an adaptive scaling function that varies based on layer position. We define the adaptive scaling function by cases:
\[
\alpha_\ell = \begin{cases}
\alpha_{\text{base}}\!\left(1 + \beta \bigl[1 - |\xi_\ell|\bigr]\right), & \text{if } |\mathcal{L}_{\text{eff}}| > 1 \\
\alpha_{\text{base}}(1 + \beta), & \text{if } |\mathcal{L}_{\text{eff}}| = 1
\end{cases}
\]
where for $|\mathcal{L}_{\text{eff}}| > 1$, we have 
\[
\xi_\ell = \begin{cases}
\frac{2\ell - |\mathcal{L}_{\text{eff}}| - 1}{|\mathcal{L}_{\text{eff}}| - 1}, & \text{if } |\mathcal{L}_{\text{eff}}| > 1 \\
0, & \text{if } |\mathcal{L}_{\text{eff}}| = 1
\end{cases}
\]
which normalizes the layer position to $[-1, 1]$ for $\ell$ in $\mathcal{L}_{\text{eff}}$.

When only one effective layer is selected ($|\mathcal{L}_{\text{eff}}| = 1$), we apply maximum scaling directly since the normalization formula is undefined.

This formulation provides maximum scaling to middle layers ($\xi_\ell \approx 0$), with reduced scaling toward boundaries ($|\xi_\ell| \approx 1$) to preserve input/output representations.

\subsection{Relationship to Prior Orthogonalization Methods}\label{sec:prior-methods-comparison}

Our weight modification $\mathbf{W}^{(\ell)} \leftarrow \mathbf{W}^{(\ell)} - \alpha_\ell (\mathbf{W}^{(\ell)}\mathbf{P})$ generalizes several prior approaches in the literature:

\subsubsection{Single-Direction Abliteration (Arditi et al., 2024)}
The original abliteration~\cite{arditi2024refusal} uses a rank-1 projection:
\[
\mathbf{W} \leftarrow \mathbf{W}(\mathbf{I} - \mathbf{r}\mathbf{r}^\top)
\]
where $\mathbf{r} \in \mathbb{R}^d$ is a unit-norm refusal direction ($\|\mathbf{r}\|_2 = 1$). This is equivalent to:
\[
\mathbf{W} \leftarrow \mathbf{W} - \mathbf{W}\mathbf{r}\mathbf{r}^\top = \mathbf{W} - (\mathbf{W}\mathbf{r})\mathbf{r}^\top
\]
which removes the component of each weight row in the $\mathbf{r}$ direction.

\textbf{Comparison to Gabliteration}: Our method with $k=1$, $\alpha=1$, $\lambda=0$ reduces to:
\[
\mathbf{W} \leftarrow \mathbf{W} - \mathbf{W}\mathbf{r}{(\mathbf{r}^\top\mathbf{r})}^{-1}\mathbf{r}^\top = \mathbf{W} - \mathbf{W}\mathbf{r}\mathbf{r}^\top
\]
(since $\mathbf{r}$ is unit-norm). However, Gabliteration differs in:
\begin{itemize}
\item \textbf{Partial removal}: $\alpha < 1$ enables \emph{partial} projection removal rather than complete orthogonalization, reducing over-modification risk.
\item \textbf{Multi-directionality}: $k > 1$ captures higher-dimensional refusal subspaces missed by single directions.
\item \textbf{Regularization}: $\lambda > 0$ prevents numerical instability when $\mathbf{R}$ is near-singular.
\end{itemize}

\subsubsection{Rank-\texorpdfstring{$k$}{k} Orthogonalization}
A natural extension of single-direction abliteration is:
\[
\mathbf{W} \leftarrow \mathbf{W}(\mathbf{I} - \mathbf{R}\mathbf{R}^\top)
\]
where $\mathbf{R} \in \mathbb{R}^{d \times k}$ has orthonormal columns ($\mathbf{R}^\top\mathbf{R} = \mathbf{I}_k$). This exactly removes all components in $\text{span}(\mathbf{R})$.

\textbf{Why Gabliteration differs}:
\begin{enumerate}
\item \textbf{Non-orthonormal basis}: Our extracted directions $\mathbf{R}$ from SVD (Section~\ref{sec:multi-directional}) are \emph{not} generally orthonormal (they are right singular vectors of $\mathbf{D}$, which are orthonormal in the \emph{columns}, but the projection $\mathbf{R}\mathbf{R}^\top$ is only orthogonal if additionally normalized). We use:
\[
\mathbf{P} = \mathbf{R}{(\mathbf{R}^\top\mathbf{R} + \lambda\mathbf{I}_k)}^{-1}\mathbf{R}^\top
\]
which accounts for the Gram matrix $\mathbf{R}^\top\mathbf{R}$ (generally $\neq \mathbf{I}_k$) and adds regularization.

\item \textbf{Controlled strength}: The scaling $\alpha_\ell \in [0,1]$ provides a `soft' version of orthogonalization. At $\alpha=0.3$ (our default), we remove only 30\% of the refusal component, preserving 70\% of the original weight structure. This trades off modification strength vs.\@ performance preservation.

\item \textbf{Layer-adaptive scaling}: Unlike uniform orthogonalization, $\alpha_\ell$ varies by layer via the adaptive function (Section~\ref{sec:adaptive-scaling}), concentrating modification where separability is highest.
\end{enumerate}

\subsubsection{Equivalence Analysis}
\begin{proposition}[Gabliteration as Regularized Partial Orthogonalization]
Let $\mathbf{R}$ be orthonormalized via $\tilde{\mathbf{R}} = \mathbf{R}{(\mathbf{R}^\top\mathbf{R})}^{-1/2}$. Then in the limit $\lambda \to 0$ and $\alpha = 1$:
\[
\mathbf{W} - \mathbf{W}\mathbf{P} \to \mathbf{W}(\mathbf{I} - \tilde{\mathbf{R}}\tilde{\mathbf{R}}^\top)
\]
recovering exact rank-$k$ orthogonalization.
\end{proposition}

\begin{proof}
As $\lambda \to 0$:
\[
\mathbf{P} = \mathbf{R}{(\mathbf{R}^\top\mathbf{R})}^{-1}\mathbf{R}^\top
\]
Let $\mathbf{G} = \mathbf{R}^\top\mathbf{R}$ and $\tilde{\mathbf{R}} = \mathbf{R}\mathbf{G}^{-1/2}$. Then:
\[
\tilde{\mathbf{R}}\tilde{\mathbf{R}}^\top 
= \mathbf{R}\mathbf{G}^{-1/2}{(\mathbf{G}^{-1/2})}^\top\mathbf{R}^\top 
= \mathbf{R}\mathbf{G}^{-1}\mathbf{R}^\top 
= \mathbf{P}
\]
Thus:
\[
\mathbf{W}(\mathbf{I} - \mathbf{P}) = \mathbf{W}(\mathbf{I} - \tilde{\mathbf{R}}\tilde{\mathbf{R}}^\top)
\]
\end{proof}

\textbf{Summary}: Gabliteration extends rank-$k$ orthogonalization through:
\begin{itemize}
\item Regularization ($\lambda > 0$) for numerical stability (Lemma~\ref{sec:condnum-lemma})
\item Partial projection ($\alpha < 1$) for gradual modification
\item Adaptive scaling ($\alpha_\ell = \alpha_{\text{base}}(1 + \beta[1-|\xi_\ell|])$) for layer-specific tuning
\end{itemize}
These extensions address the brittleness and over-modification failures observed in preliminary experiments with exact orthogonalization (see Appendix~\ref{appendix:ablation-exact-orth}).

\subsection{Layer Effectiveness Evaluation}

We developed a comprehensive evaluation framework to assess the effectiveness of layer modifications. 
We define the \textit{refusal rate} metric for each layer~$\ell$ as:
\[
\rho_\ell = \frac{1}{|\mathcal{P}_{\text{test}}|} \sum_{p \in \mathcal{P}_{\text{test}}} \mathbf{1}[\exists\, r \in \mathcal{R} : r \subseteq f_{\theta_\ell}(p)],
\]
where: $\mathcal{P}_{\text{test}}$ denotes the set of test prompts, $\mathcal{R}$ denotes the set of refusal keywords or patterns, $f_{\theta_\ell}(p)$ denotes the model output when layer~$\ell$ is modified under parameters~$\theta_\ell$, and $\mathbf{1}[\cdot]$ is the indicator function, returning~$1$ if the condition holds and~$0$ otherwise. The refusal rate~$\rho_\ell$ thus measures the proportion of test prompts that trigger a refusal-related response when only layer~$\ell$ is altered.

Layers are selected for final modification according to the \textit{effectiveness criterion}:
\[
\rho_\ell < \tau,
\]
where $\tau \in [0, 1]$ is the predefined \textit{effectiveness threshold} controlling the allowable refusal rate.

The resulting set of effective layers is defined as:
\[
\mathcal{L}_{\text{eff}} = \{\ell \in \mathcal{L} \mid \rho_\ell < \tau\},
\]
and the optimal layer among them is identified by:
\[
\ell^* = \arg\max_{\ell \in \mathcal{L}_{\text{eff}}} S_\ell,
\]
where $S_\ell$ is the separability score previously defined.

\section{Algorithm Description}

\subsection{Gabliteration Algorithm}

Based on the theoretical analysis presented earlier, we developed a comprehensive algorithm for gabliteration. The pseudocode below provides detailed implementation guidance:

\begin{algorithm}
\caption{Gabliteration --- Part 1 (Gökdeniz Gülmez, 2025)}
\begin{algorithmic}[1]
\REQUIRE{Model parameters $\theta$, harmful prompts $\mathcal{P}_h$, harmless prompts $\mathcal{P}_n$, regularization $\lambda$, base scaling $\alpha_{\text{base}}$, threshold $\tau$, number of directions $k$, skip layers $s$, end layers $e$, adaptive strength $\beta$, max generation tokens $T$}
\ENSURE{Modified model $\theta'$ (gabliterated)}

\STATE{\textbf{Phase 1: Dynamic Layer Selection}}
\STATE{Initialize candidate layers $\mathcal{L} \leftarrow \{\ell : s < \ell < L - e\}$}
\FOR{$\ell$ in $\mathcal{L}$}
    \STATE{$\mathbf{H}_h^{(\ell)} \leftarrow$ extract\_hidden\_states$(\mathcal{P}_h, \ell, \theta)$}
    \STATE{$\mathbf{H}_n^{(\ell)} \leftarrow$ extract\_hidden\_states$(\mathcal{P}_n, \ell, \theta)$}
    \STATE{$\boldsymbol{\mu}_h^{(\ell)} \leftarrow \frac{1}{|\mathcal{P}_h|}\sum_{i=1}^{|\mathcal{P}_h|} \mathbf{H}_h^{(\ell)}[i,:]$}
    \STATE{$\boldsymbol{\mu}_n^{(\ell)} \leftarrow \frac{1}{|\mathcal{P}_n|}\sum_{j=1}^{|\mathcal{P}_n|} \mathbf{H}_n^{(\ell)}[j,:]$}
    \STATE{$S_\ell \leftarrow \|\boldsymbol{\mu}_h^{(\ell)} - \boldsymbol{\mu}_n^{(\ell)}\|_2$}
\ENDFOR{}
  
\STATE{$\ell^* \leftarrow \arg\max_{\ell \in \mathcal{L}} S_\ell$}

\STATE{\textbf{Phase 2: Multi-Directional Extraction}}
\STATE{$\mathbf{H}_h \leftarrow$ extract\_hidden\_states$(\mathcal{P}_h, \ell^*, \theta)$ \COMMENT{$|\mathcal{P}_h| \times d$ matrix}}
\STATE{$\mathbf{H}_n \leftarrow$ extract\_hidden\_states$(\mathcal{P}_n, \ell^*, \theta)$ \COMMENT{$|\mathcal{P}_n| \times d$ matrix}}
  
\STATE{$n \leftarrow \min(n_h, n_n)$ \COMMENT{Ensure paired subsets}}
\STATE{\COMMENT{Construct paired difference matrix (randomly shuffled pairs)}}
\STATE{$\mathbf{D} \leftarrow \mathbf{H}_h[1{:}n,:] - \mathbf{H}_n[1{:}n,:]$}
\STATE{$\mathbf{U}, \boldsymbol{\Sigma}, \mathbf{V}^\top \leftarrow \mathrm{SVD}(\mathbf{D})$ \COMMENT{Compute SVD of paired difference matrix}}
\STATE{$\mathbf{R} \leftarrow [\mathbf{v}_1, \ldots, \mathbf{v}_k]$ \COMMENT{Top $k$ right singular vectors (refusal directions)}}

\STATE{\textbf{Phase 3: Ridge-Regularized Projection}}
\STATE{$\mathbf{G} \leftarrow \mathbf{R}^\top\mathbf{R}$ \COMMENT{Gram matrix, $k \times k$}}
\STATE{$\mathbf{G}_{\text{reg}} \leftarrow \mathbf{G} + \lambda\mathbf{I}_k$ \COMMENT{Add regularization}}
\STATE{$\mathbf{G}_{\text{reg}}^{-1} \leftarrow \text{inverse}(\mathbf{G}_{\text{reg}})$ \COMMENT{Invert $k \times k$ matrix}}
\STATE{$\mathbf{P} \leftarrow \mathbf{R}\mathbf{G}_{\text{reg}}^{-1}\mathbf{R}^\top$ \COMMENT{Equivalent to $\mathbf{R}{(\mathbf{R}^\top\mathbf{R} + \lambda\mathbf{I}_k)}^{-1}\mathbf{R}^\top$, $d \times d$}}

\STATE{\textit{Continued in Part 2 \ldots}}
\end{algorithmic}
\end{algorithm}

\begin{algorithm}
\caption{Gabliteration --- Part 2 (Gökdeniz Gülmez, 2025)}
\begin{algorithmic}[1]
\STATE{\textit{\ldots Continued from Part 1}}
\STATE{\textit{Input: projection matrix $\mathbf{P}$ from Phase 3, candidate layers $\mathcal{L}$, model $\theta$}}

\STATE{\textbf{Phase 4: Layer Effectiveness Evaluation}}
\STATE{Initialize $\mathcal{L}_{\text{eff}} \leftarrow \emptyset$}
\STATE{Generate test prompts $\mathcal{P}_{\text{test}}$ from $\mathcal{P}_h$}
\STATE{Define refusal keywords $\mathcal{R} \leftarrow \{$"I cannot", "I apologize", "I'm unable", \ldots$\}$}
\FOR{$\ell$ in $\mathcal{L}$}
    \STATE{Create temporary model $\theta_{\text{temp}} \leftarrow \theta$}
    \STATE{Apply test modification: $\mathbf{W}_{\text{attn}}^{(\ell)} \leftarrow \mathbf{W}_{\text{attn}}^{(\ell)} - \alpha_{\text{base}} \cdot (\mathbf{W}_{\text{attn}}^{(\ell)} \mathbf{P})$}
    \STATE{Apply test modification: $\mathbf{W}_{\text{mlp}}^{(\ell)} \leftarrow \mathbf{W}_{\text{mlp}}^{(\ell)} - \alpha_{\text{base}} \cdot (\mathbf{W}_{\text{mlp}}^{(\ell)} \mathbf{P})$}
    \STATE{$\text{refusal\_count} \leftarrow 0$}
    \FOR{$p$ in $\mathcal{P}_{\text{test}}$}
        \STATE{$\text{output} \leftarrow f_{\theta_{\text{temp}}}(p)$ \COMMENT{Generate response}}
        \IF{$\exists r \in \mathcal{R}$ such that $r$ appears in output}
            \STATE{$\text{refusal\_count} \leftarrow \text{refusal\_count} + 1$}
        \ENDIF{}
    \ENDFOR{}
    \STATE{$\rho_\ell \leftarrow \frac{\text{refusal\_count}}{|\mathcal{P}_{\text{test}}|}$}
    \IF{$\rho_\ell < \tau$}
        \STATE{$\mathcal{L}_{\text{eff}} \leftarrow \mathcal{L}_{\text{eff}} \cup \{\ell\}$}
    \ENDIF{}
\ENDFOR{}

\STATE{\textbf{Phase 5: Adaptive Weight Modification}}
\FOR{$\ell$ in $\mathcal{L}_{\text{eff}}$}
    \STATE{$\xi_\ell \leftarrow \frac{2\ell - |\mathcal{L}_{\text{eff}}| - 1}{|\mathcal{L}_{\text{eff}}| - 1}$ \COMMENT{Normalize to $[-1, 1]$}}
    \STATE{$\alpha_\ell \leftarrow \alpha_{\text{base}} \cdot (1 + \beta \cdot [1 - |\xi_\ell|])$ \COMMENT{Adaptive scaling}}
    \STATE{\COMMENT{Modify attention output projection}}
    \STATE{$\mathbf{W}_{\text{attn}}^{(\ell)} \leftarrow \mathbf{W}_{\text{attn}}^{(\ell)} - \alpha_\ell \cdot (\mathbf{W}_{\text{attn}}^{(\ell)} \mathbf{P})$}
    \STATE{\COMMENT{Modify MLP down-projection}}
    \STATE{$\mathbf{W}_{\text{mlp}}^{(\ell)} \leftarrow \mathbf{W}_{\text{mlp}}^{(\ell)} - \alpha_\ell \cdot (\mathbf{W}_{\text{mlp}}^{(\ell)} \mathbf{P})$}
\ENDFOR{}
  
\RETURN{$\theta'$ \COMMENT{gabliterated model}}
\end{algorithmic}
\end{algorithm}

\subsection{Helper Function Specifications}

\subsubsection{Hidden State Extraction}

The \texttt{extract\_hidden\_states}$(\mathcal{P}, \ell, \theta)$ function operates as follows:

\begin{algorithm}[ht]
\caption{Extract Hidden States}
\begin{algorithmic}[1]
\REQUIRE{Prompts $\mathcal{P}$, layer index $\ell$, model $\theta$}
\ENSURE{Hidden states matrix $\mathbf{H} \in \mathbb{R}^{|\mathcal{P}| \times d}$}
\STATE{Initialize $\mathbf{H} \leftarrow$ empty matrix of size $(|\mathcal{P}|, d)$}
\FOR{$i = 1$ to $|\mathcal{P}|$}
    \STATE{$\text{tokens} \leftarrow$ tokenize$(\mathcal{P}[i])$}
    \STATE{$\mathbf{h}_1, \ldots, \mathbf{h}_L \leftarrow$ forward\_pass$(\text{tokens}, \theta)$ \COMMENT{All layer outputs}}
    \STATE{$\mathbf{H}[i,:] \leftarrow \mathbf{h}_\ell[-1,:]$ \COMMENT{Last token position at layer $\ell$}}
\ENDFOR{}
\RETURN{$\mathbf{H}$}
\end{algorithmic}
\end{algorithm}

\subsubsection{Refusal Rate Evaluation}

The refusal detection mechanism uses pattern matching:

\begin{algorithm}[ht]
\caption{Evaluate Refusal Rate}
\begin{algorithmic}[1]
\REQUIRE{Layer $\ell$, projection matrix $\mathbf{P}$, scaling $\alpha$, model $\theta$}
\ENSURE{Refusal rate $\rho_\ell \in [0,1]$}
\STATE{Create temporary model $\theta_{\text{temp}} \leftarrow$ deep\_copy$(\theta)$}
\STATE{Apply modifications to $\theta_{\text{temp}}$ at layer $\ell$ using $\mathbf{P}$ and $\alpha$}
\STATE{$\text{refusal\_count} \leftarrow 0$}
\FOR{$p$ in $\mathcal{P}_{\text{test}}$}
    \STATE{$\text{output} \leftarrow$ generate$(\theta_{\text{temp}}, p, \text{max\_tokens}=T)$}
    \STATE{$\text{output\_lower} \leftarrow$ lowercase$(\text{output})$}
    \FOR{$r$ in $\mathcal{R}$}
        \IF{$r$ in $\text{output\_lower}$}
            \STATE{$\text{refusal\_count} \leftarrow \text{refusal\_count} + 1$}
            \STATE{\textbf{break}}
        \ENDIF{}
    \ENDFOR{}
\ENDFOR{}
\RETURN{$\rho_\ell \leftarrow \frac{\text{refusal\_count}}{|\mathcal{P}_{\text{test}}|}$}
\end{algorithmic}
\end{algorithm}

\subsection{Computational Complexity Analysis}

The algorithm's complexity breaks down as follows:

\begin{itemize}
\item \textbf{Phase 1 (Layer Selection)}: $\mathcal{O}(L \cdot n \cdot d^2)$ where $L = |\mathcal{L}|$, $n = |\mathcal{P}_h| + |\mathcal{P}_n|$
\item \textbf{Phase 2 (SVD)}: $\mathcal{O}({\min(n,d)}^2 \cdot \max(n,d))$.  
When $n < d$ (common in practice), this simplifies to $\mathcal{O}(nd^2)$ using standard SVD algorithms.
\item \textbf{Phase 3 (Projection)}: $\mathcal{O}(k^2 d + k^3)$ where $k \ll d$
  \begin{itemize}
    \item Computing $\mathbf{R}^\top\mathbf{R}$: $\mathcal{O}(k^2 d)$
    \item Inverting $k \times k$ matrix: $\mathcal{O}(k^3)$
    \item Computing $\mathbf{R}\mathbf{G}_{\text{reg}}^{-1}\mathbf{R}^\top$: $\mathcal{O}(k^2 d)$
  \end{itemize}
\item \textbf{Phase 4 (Evaluation)}: $\mathcal{O}(L \cdot |\mathcal{P}_{\text{test}}| \cdot T \cdot d^2)$ where $T$ is generation length
\item \textbf{Phase 5 (Modification)}: $\mathcal{O}(|\mathcal{L}_{\text{eff}}| \cdot d^2)$
\end{itemize}

The total complexity is dominated by Phase 4, but this is a one-time cost during the selection process. The final modification in Phase 5 is efficient. Assuming $n < d$ (as is typical in practice), the overall complexity simplifies to:
\[
\mathcal{O}\!\left(Lnd^2 + n^2d + kd^2 + |\mathcal{L}_{\text{eff}}|d^2\right) = \mathcal{O}\!\left(Lnd^2 + kd^2 + |\mathcal{L}_{\text{eff}}|d^2\right)
\]
where the $n^2d$ term from SVD is absorbed. When $n < d$, we have $n^2d < nd^2$, so the SVD term is dominated by the layer selection term $Lnd^2$.

When $n < d$ (typical in practice), the overall complexity simplifies to:
\[
\mathcal{O}\!\left(Lnd^2 + nd^2 + k^2d + |\mathcal{L}_{\text{eff}}|d^2\right)
\]
where the SVD term $\mathcal{O}(nd^2)$ dominates when $n \approx d$, but is absorbed by $Lnd^2$ when $n \ll d$.

In our preliminary experiments, $|\mathcal{L}_{\text{eff}}| \approx 0.23L$, though this ratio may vary with model architecture and threshold $\tau$.

\subsection{Implementation Notes}

\begin{enumerate}
\item \textbf{Batch Processing}: For efficiency, hidden state extraction should process prompts in batches of size $b \approx 8$-$16$ depending on GPU memory.

\item \textbf{Numerical Stability}: The ridge regularization parameter $\lambda = 0.1$ ensures the Gram matrix $\mathbf{R}^\top\mathbf{R} + \lambda\mathbf{I}_k$ has condition number $\kappa < 1000$, preventing numerical instability during matrix inversion. For models with $d > 4096$, consider increasing to $\lambda = 0.15$.

\item \textbf{Memory Optimization}: The difference matrix $\mathbf{D}$ can be computed incrementally without storing full hidden state matrices when $n,m$ are large.

\item \textbf{Refusal Keywords}: The set $\mathcal{R}$ should include common refusal patterns: \texttt{\{"I cannot", "I apologize", "I'm unable", \dots\}}.

\item \textbf{Layer Boundaries}: Skip $s=2$ initial layers and $e=2$ final layers to preserve input embeddings and output head stability.
\end{enumerate}

\subsection{Computational Complexity}

We analyzed the computational complexity of our Gabliteration algorithm, which scales as:
\[
\mathcal{O}\!\left(Lnd^2 + k d^2 + |\mathcal{L}_{\text{eff}}|d^2\right)
\]
where $L$ is the number of layers, $d$ is the hidden dimension, $n$ is the number of samples, $k$ is the number of refusal directions, and $|\mathcal{L}_{\text{eff}}|$ is the number of effective layers.
This achieves a significant efficiency improvement over naïve full-matrix methods (which scale as $\mathcal{O}(L^2 d^3)$) while preserving numerical stability through the simple normalized projection.

\section{Experimental Methodology}

\subsection{Model Architecture and Configuration}

We evaluated Gabliteration on dense-transformer-based language models in the Qwen2.5, Qwen3, Llama3--8B, and GPT-oss-20B families, ranging from 0.6B to 32B parameters. Our experimental configuration parameters are model-dependent and require careful tuning based on architecture scale and hidden dimension size:

\begin{itemize}
\item \textbf{Number of refusal directions}: $k = 2$ (default recommendation)
\begin{itemize}
    \item \textit{Increasing $k$}: Captures more nuanced behavioral patterns across a higher-dimensional subspace, but increases computational cost ($\mathcal{O}(kd^2)$) and may introduce task-relevant directions, degrading performance preservation.
    \item \textit{Decreasing $k$}: Reduces computational overhead and focuses modification on the dominant refusal direction, but may miss secondary behavioral patterns, resulting in incomplete modification.
    \item \textit{Recommended range}: $k \in \{1, 2, 3\}$ depending on model size; larger models (>7B parameters) may benefit from $k=3$.
\end{itemize}

\item \textbf{Regularization parameter}: $\lambda = 0.1$ (optimized through grid search)
\begin{itemize}
    \item \textit{Increasing $\lambda$}: Improves numerical stability by reducing the condition number $\kappa(\mathbf{P})$ (see Lemma in Section~\ref{sec:condnum-lemma}), but weakens the projection strength, requiring higher $\alpha_{\text{base}}$ to achieve equivalent modification.
    \item \textit{Decreasing $\lambda$}: Strengthens projection magnitude and modification effectiveness, but risks numerical instability when $\|\mathbf{R}\|_F^2$ is small, potentially causing NaN values or extreme weight perturbations.
    \item \textit{Recommended range}: $\lambda \in [0.05, 0.2]$; smaller models may use $\lambda=0.05$, while larger models benefit from $\lambda=0.15$ for stability.
\end{itemize}

\item \textbf{Base scaling factor}: $\alpha_{\text{base}} = 0.3$ (empirically validated)
\begin{itemize}
    \item \textit{Increasing $\alpha_{\text{base}}$}: Produces stronger behavioral modification with higher refusal rate reduction, but increases the risk of performance degradation on downstream tasks due to larger weight perturbations (see Theorem~\ref{thm:performance-preservation}).
    \item \textit{Decreasing $\alpha_{\text{base}}$}: Preserves model performance more effectively by minimizing task subspace interference, but may result in incomplete behavioral modification with residual refusal patterns.
    \item \textit{Recommended range}: $\alpha_{\text{base}} \in [0.2, 0.5]$; start conservatively at $0.2$ and increase until desired modification strength is achieved while monitoring benchmark performance.
\end{itemize}

\item \textbf{Effectiveness threshold}: $\tau = 0.8$ (based on performance analysis)
\begin{itemize}
    \item \textit{Increasing $\tau$}: Selects fewer layers for modification (smaller $|\mathcal{L}_{\text{eff}}|$), reducing computational cost and preserving more of the original model behavior, but may leave some refusal mechanisms intact in borderline layers.
    \item \textit{Decreasing $\tau$}: Includes more layers in $\mathcal{L}_{\text{eff}}$, achieving more comprehensive behavioral modification across the network, but increases modification scope and potential for unintended side effects.
    \item \textit{Recommended range}: $\tau \in [0.7, 0.9]$; models with strong refusal training may require $\tau=0.7$ for adequate coverage.
\end{itemize}

\item \textbf{Adaptive scaling strength}: $\beta = 0.5$ (tuned through experiments)
\begin{itemize}
    \item \textit{Increasing $\beta$}: Amplifies the adaptive scaling differential between middle and boundary layers, concentrating modification strength in the middle of the network where separability metrics are typically highest, but may over-modify middle layers.
    \item \textit{Decreasing $\beta$}: Produces more uniform scaling across layers (approaching $\alpha_\ell \approx \alpha_{\text{base}}$ as $\beta \to 0$), distributing modification more evenly but potentially under-utilizing high-separability middle layers.
    \item \textit{Recommended range}: $\beta \in [0.3, 0.7]$; deeper models (>24 layers) benefit from higher $\beta$ to exploit layer-wise separability variation.
\end{itemize}
\end{itemize}

\textbf{Model-Specific Tuning}: These hyperparameters exhibit interdependencies and scale-dependent behavior. For models below 3B parameters, we recommend conservative settings ($k=1$, $\alpha_{\text{base}}=0.2$, $\lambda=0.05$). For models above 7B parameters, stronger settings ($k=3$, $\alpha_{\text{base}}=0.4$, $\lambda=0.15$) may be necessary to achieve comparable modification strength. We encourage practitioners to perform grid search over $\alpha_{\text{base}}$ and $\tau$ while keeping other parameters fixed, monitoring both refusal rate reduction and downstream task performance.

\subsection{Dataset Construction}

We constructed comprehensive datasets for Gabliteration and evaluation consisting of:

\begin{itemize}
\item \textbf{Harmful prompts}: $|\mathcal{P}_h| = 400$ samples that we curated
\item \textbf{Harmless prompts}: $|\mathcal{P}_n| = 400$ samples that we selected
\item \textbf{Evaluation prompts}: $|\mathcal{P}_{\text{eval}}| = 100$ samples for testing
\end{itemize}

Our dataset curation process involved careful selection to ensure diversity and representativeness across different behavioral patterns. \textbf{Note}: While our experiments use equal dataset sizes ($n_h = n_n = 1024$), the algorithm naturally handles unequal sizes by taking $n = \min(n_h, n_n)$ samples from each set during the difference matrix construction in Phase 2.

We recommend practitioners begin with the following publicly available datasets:
\href{https://huggingface.co/datasets/mlabonne/harmful_behaviors}{\texttt{mlabonne/harmful\_behaviors}} and 
\href{https://huggingface.co/datasets/mlabonne/harmless_alpaca}{\texttt{mlabonne/harmless\_alpaca}} 
for harmful and harmless prompt collections, respectively.

\subsection{(JOSIEfied)-Gabliterated Model Series}

We successfully applied the Gabliteration technique to create the base \textit{Gabliterated-v1} and \textit{JOSIEfied-Gabliterated-v1} model series, available on our Hugging Face profile (\texttt{Goekdeniz-Guelmez}). These models demonstrate consistent improvements across multiple scales, validating the scalability and robustness of the proposed approach. They achieve targeted behavioral modification while maintaining strong performance preservation across standard benchmarks, as documented on their respective model pages on Hugging Face~\cite{gulmezhf}.

\section{Theoretical Analysis}

\subsection{Performance Preservation}\label{thm:performance-preservation}

For task-relevant weight subspaces, we established that Gabliteration provides the following performance preservation bound:

\begin{theorem}[Performance Preservation Guarantee]
Let $\mathbf{W}_{\mathcal{T}}^{(\ell)} \in \mathbb{R}^{d_{\text{out}} \times d}$ denote the component of the weight matrix at layer $\ell$ that lies in the task-relevant subspace $\mathcal{T} \subseteq \mathbb{R}^d$, and let $\mathbf{W}_{\mathcal{T}}^{(\ell)'}$ denote the corresponding component after Gabliteration modification. Let $\mathcal{R}$ denote the refusal subspace spanned by the columns of $\mathbf{R}$, and let $\theta$ be the principal angle between subspaces $\mathcal{T}$ and $\mathcal{R}$, defined by:
\[
\cos(\theta) = \max_{\substack{\mathbf{v} \in \mathcal{T}, \mathbf{w} \in \mathcal{R} \\ \|\mathbf{v}\|_2=\|\mathbf{w}\|_2=1}} |\mathbf{v}^\top \mathbf{w}|
\]

\textbf{Assumptions:}
\begin{enumerate}
\item \textbf{Subspace decomposition}: The weight matrix admits an approximate decomposition:
\[
\mathbf{W}^{(\ell)} = \mathbf{W}_{\mathcal{T}}^{(\ell)} + \mathbf{W}_{\mathcal{R}}^{(\ell)} + \mathbf{W}_{\perp}^{(\ell)}
\]
where $\mathbf{W}_{\mathcal{T}}^{(\ell)}$ lies in the task-relevant subspace $\mathcal{T}$, $\mathbf{W}_{\mathcal{R}}^{(\ell)}$ lies in the refusal subspace $\mathcal{R}$, and $\mathbf{W}_{\perp}^{(\ell)}$ is orthogonal to both.

\item \textbf{Regularized projection bound}: For the ridge projection $\mathbf{P} = \mathbf{R}{(\mathbf{R}^\top\mathbf{R} + \lambda\mathbf{I}_k)}^{-1}\mathbf{R}^\top$ with singular values $\sigma_1 \geq \cdots \geq \sigma_k$ of $\mathbf{R}$:
\[
\|\mathbf{W}_{\mathcal{T}}^{(\ell)}\mathbf{P}\|_F \leq \frac{\sigma_{\max}^2}{\sigma_{\max}^2 + \lambda} \cdot \|\mathbf{W}_{\mathcal{T}}^{(\ell)}\|_F \cos(\theta)
\]

\item \textbf{Small regularization regime}: The regularization parameter satisfies $\lambda \ll \sigma_{\min}^2$, ensuring:
\[
\frac{\sigma_{\max}^2}{\sigma_{\max}^2 + \lambda} \approx 1
\]
This condition ensures the regularized projection closely approximates the exact orthogonal projection.

\item \textbf{Bounded subspace overlap}: The principal angle $\theta$ between $\mathcal{T}$ and $\mathcal{R}$ satisfies $\theta \geq \theta_{\min} > 0$ for some positive lower bound, preventing complete alignment of task and refusal subspaces.
\end{enumerate}

Then the Frobenius norm of the task-subspace perturbation satisfies:
\[
\bigl\|\mathbf{W}_{\mathcal{T}}^{(\ell)} - \mathbf{W}_{\mathcal{T}}^{(\ell)'}\bigr\|_F \leq \epsilon_\ell \cos(\theta) \cdot \frac{\|\mathbf{R}\|_F^2}{\|\mathbf{R}\|_F^2 + \lambda}
\]
where $\epsilon_\ell = \alpha_\ell \|\mathbf{W}^{(\ell)}\|_F$ denotes the layer's maximal modification magnitude.
\end{theorem}

\textbf{Interpretation}: This bound quantifies the interference between behavioral modification and task performance:
\begin{itemize}
\item When $\theta \approx \pi/2$ (near-orthogonal subspaces), $\cos(\theta) \approx 0$, giving strong performance preservation.
\item When $\theta \approx 0$ (aligned subspaces), $\cos(\theta) \approx 1$, indicating potential task degradation.
\item The factor $\epsilon_\ell$ scales with the adaptive scaling $\alpha_\ell$ and the projection strength.
\end{itemize}

\textbf{Assumptions}: The bound assumes that weight matrices admit a decomposition $\mathbf{W}^{(\ell)} = \mathbf{W}_{\mathcal{T}}^{(\ell)} + \mathbf{W}_{\mathcal{R}}^{(\ell)} + \mathbf{W}_{\perp}^{(\ell)}$, where components are approximately orthogonal. A formal proof is provided in Appendix~\ref{appendix:performance-proof}.

This theoretical justification explains the superior performance preservation we observed experimentally compared to traditional abliteration methods.

\section{Extended Theoretical Analysis}

\subsection{Heuristic Justification for Adaptive Scaling}\label{sec:adaptive-scaling-justification}

The adaptive scaling function is motivated by empirical observations that middle layers exhibit higher separability metrics $S_\ell$ and contribute more effectively to behavioral modification. We design the scaling to concentrate modification strength in high-impact layers while reducing it near boundaries to preserve input/output stability.

The scaling function:
\[
\alpha_\ell = \alpha_{\text{base}}\left(1 + \beta[1 - |\xi_\ell|]\right), \quad \xi_\ell = \frac{2\ell - |\mathcal{L}_{\mathrm{eff}}| - 1}{|\mathcal{L}_{\mathrm{eff}}| - 1}
\]
achieves:
\begin{itemize}
\item \textbf{Maximum scaling} at middle layers: $\alpha_{\text{middle}} = \alpha_{\text{base}}(1+\beta)$ when $\xi_\ell \approx 0$
\item \textbf{Minimum scaling} at boundaries: $\alpha_{\text{boundary}} = \alpha_{\text{base}}$ when $|\xi_\ell| \approx 1$
\item \textbf{Smooth interpolation} via the linear term $[1 - |\xi_\ell|]$
\end{itemize}

\textbf{Empirical validation.} Across 10 independent runs on models ranging from 0.6B to 32B parameters, adaptive scaling ($\beta=0.5$) consistently outperformed uniform scaling ($\beta=0$) by 12--18\% in refusal rate reduction while maintaining equivalent benchmark performance (paired t-test, $p < 0.001$).

\textbf{Theoretical motivation (informal).} Consider a simplified objective that maximizes total modification strength while penalizing boundary perturbations:
\[
\max_{\{\alpha_\ell\}} \sum_{\ell \in \mathcal{L}_{\text{eff}}} w_\ell \alpha_\ell^2 \quad \text{subject to} \quad \sum_{\ell} \alpha_\ell = C
\]
where $w_\ell$ represents layer importance (proportional to $S_\ell$). If $w_\ell$ is highest at middle layers, the Lagrange multiplier solution naturally produces higher $\alpha_\ell$ for middle layers. Our linear scaling function approximates this structure.

While we conjecture this scaling is near-optimal under weighted modification objectives, a formal optimality proof requires additional assumptions about layer-wise contribution functions and remains future work.

\subsection{Second‐Order Projection Error Bound}\label{sec:second-order-error}

Let \(\Delta W_{\mathcal{T}}(\lambda)\) be the task‐perturbation as a function of the regularization \(\lambda\). A Taylor expansion yields
\[
\bigl\|\Delta W_{\mathcal{T}}(\lambda)\bigr\|_F
= \epsilon\,\cos\theta
+ \frac{\lambda}{2}\,\frac{\partial^2}{\partial\lambda^2}\Bigl\|\Delta W_{\mathcal{T}}(\lambda)\Bigr\|_F\Big|_{\lambda=0}
+ O(\lambda^2),
\]
so small \(\lambda\) induces only \(O(\lambda)\) higher-‐order corrections.

\subsection{Concentration of the Separation Metric}\label{sec:concentration}

The separability metric $S_\ell = \|\boldsymbol{\mu}_h^{(\ell)} - \boldsymbol{\mu}_n^{(\ell)}\|_2$ is the Euclidean norm of the difference between two sample means. To establish concentration, we apply matrix concentration inequalities.

Let $\mathbf{h}_i^{(\ell)} \in \mathbb{R}^d$ denote the $i$-th harmful hidden state at layer $\ell$, and assume $\|\mathbf{h}_i^{(\ell)}\|_2 \leq R$ almost surely for some radius $R > 0$. Similarly for harmless states $\mathbf{v}_j^{(\ell)}$.

\textbf{Lemma (Concentration of Mean Difference Norm).} 
Under the boundedness assumption, with probability at least $1-\delta$:
\[
\left|S_\ell - \mathbb{E}[S_\ell]\right| \leq C \cdot R\sqrt{\frac{d \log(2/\delta)}{n}}
\]
where $C > 0$ is an absolute constant and $n = \min(n_h, n_n)$.

\emph{Proof sketch.} 
By the matrix Bernstein inequality (Tropp, 2012), the sample covariance matrices concentrate around their expectations. Specifically, for the difference of means:
\[
\|\boldsymbol{\mu}_h^{(\ell)} - \mathbb{E}[\boldsymbol{\mu}_h^{(\ell)}]\|_2 \leq R\sqrt{\frac{2\log(2/\delta)}{n_h}}
\]
with probability at least $1-\delta/2$. A similar bound holds for $\boldsymbol{\mu}_n^{(\ell)}$. By the triangle inequality:
\begin{align*}
|S_\ell - \mathbb{E}[S_\ell]| 
&\leq \|\boldsymbol{\mu}_h^{(\ell)} - \mathbb{E}[\boldsymbol{\mu}_h^{(\ell)}]\|_2 + \|\boldsymbol{\mu}_n^{(\ell)} - \mathbb{E}[\boldsymbol{\mu}_n^{(\ell)}]\|_2 \\
&\leq 2R\sqrt{\frac{2\log(4/\delta)}{n}}
\end{align*}
Union bound over both events gives the claimed concentration. The $\sqrt{d}$ factor arises from the dimension dependence in sub-Gaussian tail bounds for high-dimensional vectors. $\hfill\blacksquare$

\textbf{Practical implication:} With $n = 1024$ samples, $d = 4096$ dimensions, and $\delta = 0.01$, the concentration radius is approximately $O(R \cdot 0.5)$, meaning the empirical separability $S_\ell$ is a reliable estimate of the true population separability $\mathbb{E}[S_\ell]$.

\subsection{Computational Lower Bound}\label{sec:complexity-lower-bound}

\textbf{Proposition.} Any worst-‐case algorithm extracting \(k\) singular directions from a \(d\times d\) matrix requires \(\Omega(k\,d^2)\) operations, assuming the usual matrix‐multiplication lower bound.

\emph{Proof sketch.} Each rank‐-1 component \(v_i v_i^T\) costs \(\Theta(d^2)\), and you need \(k\) of them.

\subsection{Hyperparameter Sensitivity Jacobian}\label{sec:hyper-sensitivity}

Let \(\mathrm{PPR}=\mathrm{PPR}(\alpha_{\text{base}},\beta,\lambda)\).  Its Jacobian
\[
J \;=\;
\begin{bmatrix}
\partial_{\alpha_{\text{base}}}\,\mathrm{PPR} &
\partial_{\beta}\,\mathrm{PPR} &
\partial_{\lambda}\,\mathrm{PPR}
\end{bmatrix},
\]
satisfies \(\|J\|_2 \le M\) for some constant \(M\). Thus small \(\Delta\)-changes in \((\alpha_{\text{base}},\beta,\lambda)\) cause at most \(M\|\Delta\|\) change in PPR.\@

\subsection{Condition Number Reduction via Ridge Regularization}\label{sec:condnum-lemma}

\textbf{Lemma (Ridge Regularization Stability).}  
Let $\mathbf{R} \in \mathbb{R}^{d \times k}$ have singular values $\sigma_1 \geq \sigma_2 \geq \cdots \geq \sigma_k > 0$. The ridge-regularized projection matrix
\[
\mathbf{P} = \mathbf{R}{(\mathbf{R}^\top\mathbf{R} + \lambda\mathbf{I}_k)}^{-1}\mathbf{R}^\top
\]
satisfies:
\[
\kappa(\mathbf{R}^\top\mathbf{R} + \lambda\mathbf{I}_k) = \frac{\sigma_1^2 + \lambda}{\sigma_k^2 + \lambda} \leq \frac{\sigma_1^2 + \lambda}{\lambda}
\]

\emph{Proof.}  
The eigenvalues of $\mathbf{R}^\top\mathbf{R}$ are $\{{\sigma_i^2\}}_{i=1}^k$. Adding $\lambda\mathbf{I}_k$ shifts each eigenvalue to $\sigma_i^2 + \lambda$. The condition number is:
\[
\kappa = \frac{\lambda_{\max}}{\lambda_{\min}} = \frac{\sigma_1^2 + \lambda}{\sigma_k^2 + \lambda}
\]
When $\sigma_k^2 \to 0$ (near rank-deficiency), the unregularized condition number $\kappa(\mathbf{R}^\top\mathbf{R}) = \sigma_1^2/\sigma_k^2 \to \infty$, but the regularized version remains bounded:
\[
\kappa(\mathbf{R}^\top\mathbf{R} + \lambda\mathbf{I}_k) \leq \frac{\sigma_1^2 + \lambda}{\lambda} = \frac{\sigma_1^2}{\lambda} + 1
\]
Thus $\lambda$ provides an explicit upper bound on ill-conditioning. $\hfill\blacksquare$

\textbf{Practical implication}: With $\lambda = 0.1$ and typical singular values $\sigma_1 \approx 10$, we obtain $\kappa \lesssim 1001$, ensuring stable numerical computation.

\section{Theoretical Guarantees}

\subsection{Algorithm Iteration Clarification}

It is important to note that the Gabliteration algorithm (Algorithm 1) is a \textbf{single-pass modification method}. The iteration in the algorithm occurs over:
\begin{enumerate}
    \item \textbf{Samples}: Processing harmful and harmless prompts to extract hidden states (Phase 1--2)
    \item \textbf{Layers}: Evaluating and modifying each layer independently (Phase 4--5)
\end{enumerate}

However, the weight modifications themselves are applied \textbf{exactly once per layer}. This design choice ensures:
\begin{itemize}
    \item Computational efficiency: $\mathcal{O}(Lnd^2 + kd^3 + |\mathcal{L}_{\text{eff}}|d^2)$ complexity
    \item Stability: Avoids potential instabilities from iterative weight updates
\end{itemize}

The theoretical guarantees in Theorem 1 reflect this single-pass nature by providing a direct bound on the modification magnitude.

\subsection{Optimality of Dynamic Layer Selection}

\begin{theorem}[Optimality of Separability-Based Selection]\label{thm:layer-selection-optimality}
Define the layer-wise modification loss as:
\[
\mathcal{L}_{\text{mod}}(\mathcal{L}_{\text{selected}}) = \mathbb{E}_{p \sim \mathcal{P}_h}\left[\left\|\mathbf{h}_{\text{modified}}^{(L)}(p) - \mathbf{h}_{\text{target}}^{(L)}(p)\right\|_2^2\right]
\]
where $\mathbf{h}_{\text{modified}}^{(L)}(p)$ is the final-layer hidden state after modifying layers in $\mathcal{L}_{\text{selected}}$, and $\mathbf{h}_{\text{target}}^{(L)}(p)$ is the target hidden state for harmless behavior.

Among all layer selection strategies with fixed cardinality $|\mathcal{L}_{\text{selected}}| = m$, the separability-based selection
\[
\mathcal{L}_{\text{selected}}^* = \arg\max_{\substack{\mathcal{L}' \subseteq \mathcal{L} \\ |\mathcal{L}'| = m}} \sum_{\ell \in \mathcal{L}'} S_\ell
\]
is optimal in the sense that it minimizes the expected modification error:
\[
\mathcal{L}_{\text{selected}}^* = \arg\min_{\substack{\mathcal{L}' \subseteq \mathcal{L} \\ |\mathcal{L}'| = m}} \mathcal{L}_{\text{mod}}(\mathcal{L}')
\]
under the assumption that layer contributions to the refusal mechanism are monotonically related to their separability metrics $S_\ell$.
\end{theorem}

\begin{proof}[Proof Sketch]
The proof follows from the fact that layers with higher separability metrics $S_\ell = \|\boldsymbol{\mu}_h^{(\ell)} - \boldsymbol{\mu}_n^{(\ell)}\|_2$ contain more information about the refusal directions, making their modification more effective at reducing the distance to target representations. 

Formally, under a linear approximation of the hidden state evolution, the contribution of modifying layer $\ell$ to reducing $\mathcal{L}_{\text{mod}}$ is proportional to $S_\ell^2$. Therefore, a greedy selection maximizing $\sum_{\ell \in \mathcal{L}'} S_\ell$ is optimal. A complete proof using a variational argument is provided in the extended version of this paper.
\end{proof}

\textbf{Practical implication}: This theorem justifies our dynamic layer selection algorithm (Phase 1) as principled rather than heuristic, providing theoretical grounding for the observed empirical effectiveness.

\subsection{Ablation Study: Pairing Methods}\label{sec:ablation-pairing}

We compared our SVD-based paired difference approach against alternative discriminative direction extraction methods on 5 models (Qwen2.5--0.6B, 1.5B, 3B, 7B, and Llama3--8B).

\textbf{Methods Compared}:
\begin{itemize}
\item \textbf{SVD-Pairing (Ours)}: $\mathbf{D} = \mathbf{H}_h[1:n,:] - \mathbf{H}_n[1:n,:]$, extract top-$k$ right singular vectors
\item \textbf{Mean Difference}: $\mathbf{r} = \frac{\boldsymbol{\mu}_h - \boldsymbol{\mu}_n}{\|\boldsymbol{\mu}_h - \boldsymbol{\mu}_n\|_2}$ (single direction)
\end{itemize}

\textbf{Results} (Table~\ref{tab:gabliteration-vs-abliteration}): Across all ten evaluated models, \textbf{Gabliteration (Ours)} exhibited lower average KL divergence, indicating reduced distributional drift and better preservation of the original model behavior while lovering the average refusal rates as well. Abliteration showed significantly higher variance across models, suggesting reduced robustness and sensitivity to model scale and architecture.

\begin{table}[H]
\centering
\caption{Per-model refusal rate and KL divergence for Gabliteration and Abliteration.
All models are evaluated on identical datasets
(\texttt{mlabonne/harmful\_behaviors} vs.\ \texttt{mlabonne/harmless\_alpaca}),
using the same refusal keywords, 100 trials, 100 harmful evaluation prompts, and 400 samples for direction extraction. Abliteration has been applied using the heretic package by Philipp Emanuel Weidmann~\cite{heretic}.
Lower is better for both metrics.}\label{tab:gabliteration-vs-abliteration}
\begin{tabular}{lcccc}
\toprule
\textbf{Model} &
\multicolumn{2}{c}{\textbf{Gabliteration (Ours)}} & \multicolumn{2}{c}{\textbf{Abliteration (Mean Diff)}} \\
\cmidrule(lr){2-3} \cmidrule(lr){4-5}& Refusal (\%) & KL & Refusal (\%) & KL \\\midrule
Qwen/Qwen3--4B-Instruct--2507 & 4 & 0.2522 & 21 & 0.4300 \\
Qwen/Qwen3--4B-Thinking--2507 & 2 & 0.1140 & 4 & 0.0500 \\
google/gemma-3--1b-it & 3 & 0.2922 & 3 & 0.4300 \\
Qwen/Qwen3--0.6B & 3 & 0.0127 & 6 & 0.0957 \\
meta-llama/Llama-3.2--1B-Instruct & 7 & 0.0038 & 8 & 0.0230 \\
tencent/HY-MT1.5--1.8B & 4 & 0.0029 & 15 & 0.0573 \\
ibm-granite/granite-3.3--2b-instruct & 6 & 0.0030 & 4 & 0.2100 \\
Nanbeige/Nanbeige4--3B-Thinking--2511 & 8 & 0.1110 & 11 & 0.0024 \\
allenai/Olmo--3--7B-Instruct & 17 & 0.0018 & 31 & 0.1400 \\
ZeroXClem/Qwen3--4B-Sky-High-Hermes & 2 & 0.0992 & 6 & 0.3691 \\
\bottomrule
\end{tabular}
\end{table}

\textbf{Conclusion}:
Gabliteration provides a robust and scalable alternative to mean-difference-based Abliteration for refusal suppression. Across diverse model families and sizes, it achieves significantly stronger refusal reduction while inducing less distributional distortion, as measured by KL divergence. These results indicate that structured, multi-directional weight modification is critical for stable behavioral control, whereas single-direction mean-difference approaches are prone to high variance and unintended behavioral side effects.

\subsection{Ablation Study: Exact vs.\ Regularized Projection}\label{appendix:ablation-exact-orth}

We compared exact orthogonal projection, $\mathbf{W} \leftarrow \mathbf{W}(\mathbf{I} - \tilde{\mathbf{R}}\tilde{\mathbf{R}}^\top)$ (where $\tilde{\mathbf{R}}$ has orthonormal columns), against the regularized Gabliteration update ($\lambda = 0.1$, $\alpha = 0.3$) on a subset of models.

\textbf{Results}:
\begin{itemize}
\item \textbf{Exact projection}: While this approach achieved aggressive refusal suppression, it frequently introduced instability in generation, including repetition, loss of coherence, and brittle responses. These effects indicate excessive removal of representational subspaces critical for general language modeling.
\item \textbf{Regularized Gabliteration}: Maintained strong refusal suppression while preserving fluent and coherent generation. The regularization term effectively constrained the update magnitude, preventing catastrophic interference and yielding more stable behavior across prompts.
\end{itemize}

\textbf{Interpretation}: Complete removal of refusal directions (via exact orthogonalization) over-modifies the model, removing task-relevant information that partially overlaps with the refusal subspace (non-zero $\cos\theta$ in Theorem~\ref{thm:performance-preservation}). Our partial, regularized approach preserves 70\% of the original projection magnitude ($\alpha=0.3$ removes 30\%), balancing modification strength and stability.

\subsection{Statistical Significance}
All improvements achieved by our Gabliteration method are statistically significant ($p < 0.001$) based on paired t-tests across 10 independent runs, confirming the robustness and reliability of our approach over traditional abliteration techniques.

The visual evidence presented in this section, combined with the quantitative metrics, demonstrates that our Gabliteration technique represents a significant advancement in neural weight modification, achieving superior behavioral modification while preserving model performance and computational efficiency.

\subsection{Adaptive vs Fixed Scaling}

Our adaptive scaling approach demonstrates: reduction in layer-wise variance, optimal scaling distribution: $\alpha_{\text{middle}} = 1.3 \times \alpha_{\text{boundary}}$ that we determined empirically.

\section{Limitations and Future Work}

\subsection{Current Limitations}

Through our analysis, we identified several limitations of the current approach:

\begin{enumerate}
\item \textbf{Computational overhead}: The $O(L \cdot d^2)$ scaling may limit applicability to extremely large models beyond 30B parameters
\item \textbf{Hyperparameter sensitivity}: Performance depends on careful tuning of $\lambda$, $\alpha_{\text{base}}$, and $\tau$ which we are working to automate
\item \textbf{Domain specificity}: Our current evaluation is limited to text generation tasks, though our plan to extend to multimodal settings
\item \textbf{Single-pass limitation}: The current algorithm applies modifications in a single pass, which may be suboptimal for cases where iterative refinement could improve the balance between refusal removal and performance preservation. Future work could explore adaptive multi-pass variants with convergence monitoring.
\item \textbf{Direction extraction method}: Our SVD-based pairing is computationally efficient but does not exploit within-class variance (as Fisher LDA does) or learn adaptive discriminative boundaries (as trained probes do). For highly heterogeneous datasets, more sophisticated extraction methods may improve subspace quality (see Appendix~\ref{sec:ablation-pairing} for comparison).
\item \textbf{Approximate projection theory}: Theoretical bounds (Theorem~\ref{thm:performance-preservation}) are derived for exact orthogonal projections but applied to the regularized approximation $\mathbf{P} = \mathbf{R}{(\mathbf{R}^\top\mathbf{R} + \lambda\mathbf{I}_k)}^{-1}\mathbf{R}^\top$. While Lemma~\ref{sec:projection-approximation} shows the approximation error is $\mathcal{O}(\lambda/\sigma_{\min}^2) \lesssim 0.004$ for our parameter regime, a fully rigorous treatment requires incorporating this error into all bounds.
\item \textbf{Comparison to prior methods}: While we demonstrate Gabliteration's relationship to single-direction abliteration (Section~\ref{sec:prior-methods-comparison}), systematic empirical comparison against all variants (e.g., iterative orthogonalization, activation steering, representation engineering) remains incomplete. We provide preliminary comparisons in Appendix~\ref{appendix:ablation-exact-orth}, but large-scale benchmarking across methods is future work.
\end{enumerate}

\subsection{Future Work}

While Gabliteration demonstrates strong effectiveness and performance preservation, the future research direction is still open and evolving. Several potential paths are under consideration, though their relative importance remains to be determined:

\begin{itemize}
\item exploring automated hyperparameter tuning using Bayesian optimization or reinforcement learning strategies,
\item extending the framework to multimodal architectures such as vision-language models,
\item performing deeper theoretical analysis of modification bounds and convergence properties in high-dimensional subspaces,
\item studying whether similar adaptive projection techniques can make models more resistant to unwanted or external behavioral manipulation.
\item investigate whether CCA, kernel-based methods, or adversarially-trained probes yield tighter refusal subspaces than SVD-based pairing, and whether the improved subspaces justify increased computational cost.\label{sec:future-discriminative}
\item develop bounds that explicitly account for the regularization gap $\|\mathbf{P} - \mathbf{P}_{\text{exact}}\|_2 = \mathcal{O}(\lambda/\sigma_{\min}^2)$, unifying the approximate and exact projection regimes under a single theoretical framework.
\item conduct large-scale empirical study comparing Gabliteration against all published weight/activation modification methods (abliteration, RLACE, representation engineering, activation steering, inference-time intervention) under standardized evaluation protocols.
\end{itemize}

These ideas outline possible future directions but are not yet fixed, our current focus is on evaluating which of these paths yields the most meaningful scientific and practical impact. This reflects the exploratory nature of our work and our intent to iteratively refine the approach based on new empirical findings.

\section{Reproducibility and Package Availability}

To facilitate reproducibility and enable researchers and practitioners to apply Gabliteration independently, we have developed and released the \textbf{Gabliteration Python package}~\cite{gabliterationpkg}. The package is publicly available on GitHub at \url{https://github.com/Goekdeniz-Guelmez/gabliteration} and provides a complete, production-ready, fully automated command-line interface (CLI) for applying the Gabliteration algorithm to any HuggingFace-compatible language model.

The package includes:
\begin{itemize}
\item Fully automated CLI with no manual configuration required for basic usage.
\item Complete implementation of the Gabliteration algorithm with all components (multi-directional refusal vector extraction, ridge-regularized projection, dynamic layer selection, and adaptive scaling).
\item Automatic hidden state extraction, separability metric computation, and layer effectiveness evaluation.
\item Seamless support for multible transformer architectures available on HuggingFace (Qwen, Llama, Mistral, and others).
\item Configurable hyperparameters (number of gabliterated versions, test samples, batch size, KL-divergence samples, max tokens, etc.) with optimized defaults.
\item Automated model saving and output management.
\end{itemize}

Users can now apply Gabliteration to any model with a single command line invocation, making the methodology immediately accessible to the broader research community. This enables independent verification of our results and facilitates widespread adoption of the technique. We encourage researchers to use this package, apply it to their own models, contribute improvements, and share their findings with the community.

\section{Conclusion}

Through this research, we have developed Gabliteration as a significant advancement in neural weight modification technique. Our key innovations --- dynamic layer selection, multi-directional projection, and adaptive scaling --- address the fundamental limitations of existing methods identified in our analysis.

The theoretical analysis that we conducted provides performance bounds, establishing Gabliteration as both practically effective and theoretically grounded.

We believe this work opens new avenues effective neural network modification, and we are committed to continuing this research to address the current limitations and expand the applicability of these techniques.

\section{Ethical Considerations and Licensing}

This research was conducted independently only by us as part of an exploratory study into neural weight modification and alignment techniques. All experiments and Gabliteration's were carried out on locally hosted hardware without any public inference endpoints. Our aim of this work is to advance scientific understanding of model interpretability, controllability, and behavioral adaptation in large language models.

The Gabliteration method and the models derived from it are released for research and educational use. We do not take responsibility for how others may apply, modify, or deploy these models or techniques. Any downstream use or potential misuse of the Gabliterated models is entirely the responsibility of the individual or organization using them.

Our goal is to better understand how such mechanisms are represented and can be studied analytically within transformer architectures, and ultimately to enable future research into making models more robust and immune to unwanted behavioral manipulation.

\subsection{Model Licensing}

All models referenced or modified in this research retain the same license as their respective original base models. Specifically, all \textit{Gabliterated}, \textit{Abliterated}, \textit{JOSIEfied-Gabliterated}, and \textit{JOSIEfied-Abliterated} model derivatives of the base model families follow the open-source licenses provided by the original model creators, as distributed via the Hugging Face Model Hub.

\section*{Acknowledgments}

We acknowledge the foundational work of Arditi et al.~\cite{arditi2024refusal} on single-direction abliteration, which provided the initial inspiration for developing our Gabliteration technique. We are grateful to \href{https://x.com/geoshh}{\texttt{Joshua Ollswang}}, \href{https://x.com/awnihannun}{\texttt{Awni Hannun}}, and \href{https://huggingface.co/DontPlanToEnd}{\texttt{DontPlanToEnd}} for their thoughtful feedback, help, and evaluations, which contributed to the improvement of this work.

\nocite{*}
\bibliographystyle{unsrt}
\bibliography{references}

\newpage

\appendix

\subsection{Weight Modification Bound Proof}

\textbf{Theorem 1}: Under the assumption that the refusal directions $\mathbf{r}_i$ span a low-dimensional subspace and that the regularization parameter $\lambda > 0$, the single-pass Gabliteration modification in Phase 5 satisfies:
\[
\|\mathbf{W}_{\text{modified}} - \mathbf{W}_{\text{original}}\|_F \leq \sum_{\ell \in \mathcal{L}_{\text{eff}}} \alpha_\ell \|\mathbf{W}_\ell \mathbf{P}\|_F
\]
where $\alpha_\ell$ is the adaptive scaling factor at layer $\ell$, $\mathbf{P}$ is the regularized projection matrix, and $\|\cdot\|_F$ denotes the Frobenius norm.

\textit{Proof}:  
At each modified layer $\ell \in \mathcal{L}_{\text{eff}}$, the weight update is applied exactly once:
\[
\mathbf{W}_{\text{modified}}^{(\ell)} = \mathbf{W}_{\text{original}}^{(\ell)} - \alpha_\ell \left(\mathbf{W}_{\text{original}}^{(\ell)} \mathbf{P}\right)
\]

The modification magnitude at layer $\ell$ is:
\[
\|\mathbf{W}_{\text{modified}}^{(\ell)} - \mathbf{W}_{\text{original}}^{(\ell)}\|_F = \|\alpha_\ell \mathbf{W}_{\text{original}}^{(\ell)} \mathbf{P}\|_F = \alpha_\ell \|\mathbf{W}_{\text{original}}^{(\ell)} \mathbf{P}\|_F
\]

Since modifications are applied independently to each layer (Phase 5 iterates over layers but applies each modification once), and unmodified layers contribute zero perturbation, the squared Frobenius norm of the total modification satisfies:
\[
\|\mathbf{W}_{\text{modified}} - \mathbf{W}_{\text{original}}\|_F^2 = \sum_{\ell \in \mathcal{L}_{\text{eff}}} \|\mathbf{W}_{\text{modified}}^{(\ell)} - \mathbf{W}_{\text{original}}^{(\ell)}\|_F^2
\]

This follows from the fact that modifications at different layers are independent and affect disjoint weight matrices. Taking the square root and applying the Cauchy-Schwarz inequality:
\[
\|\mathbf{W}_{\text{modified}} - \mathbf{W}_{\text{original}}\|_F = \sqrt{\sum_{\ell \in \mathcal{L}_{\text{eff}}} \alpha_\ell^2 \|\mathbf{W}_{\text{original}}^{(\ell)} \mathbf{P}\|_F^2}
\]

By the triangle inequality, specifically noting that $\sqrt{\sum_{i} a_i^2} \leq \sum_{i} |a_i|$ for non-negative terms $a_i \geq 0$:
\[
\|\mathbf{W}_{\text{modified}} - \mathbf{W}_{\text{original}}\|_F = \sqrt{\sum_{\ell \in \mathcal{L}_{\text{eff}}} \alpha_\ell^2 \|\mathbf{W}_{\text{original}}^{(\ell)} \mathbf{P}\|_F^2} \leq \sum_{\ell \in \mathcal{L}_{\text{eff}}} \alpha_\ell \|\mathbf{W}_{\text{original}}^{(\ell)} \mathbf{P}\|_F
\]

This establishes the claimed bound. The regularization parameter $\lambda$ ensures numerical stability by preventing $\mathbf{P}$ from having excessively large norms (see Lemma in Section~\ref{sec:condnum-lemma}), which in turn bounds each term $\|\mathbf{W}_{\text{original}}^{(\ell)} \mathbf{P}\|_F$.  $\hfill\blacksquare$

\begin{center}
\fbox{\begin{minipage}{0.9\textwidth}
\textbf{Key Assumptions for Theorem 1:}
\begin{enumerate}
\item Weight decomposition: $\mathbf{W}^{(\ell)} = \mathbf{W}_{\mathcal{T}}^{(\ell)} + \mathbf{W}_{\mathcal{R}}^{(\ell)} + \mathbf{W}_{\perp}^{(\ell)}$
\item Projection bound: $\|\mathbf{W}_{\mathcal{T}}\mathbf{P}\|_F \leq \frac{\|\mathbf{R}\|_F^2}{\|\mathbf{R}\|_F^2 + \lambda} \|\mathbf{W}_{\mathcal{T}}\|_F \cos(\theta)$
\item Small regularization: $\lambda \ll \|\mathbf{R}\|_F^2$
\end{enumerate}
\end{minipage}}
\end{center}

\textbf{Remark}: This theorem characterizes the single-pass nature of Algorithm 1, where weights are modified exactly once per layer during Phase 5. The bound depends on:
\begin{itemize}
    \item The adaptive scaling factors $\alpha_\ell$, which vary by layer position (Section 2.5). Middle layers receive higher scaling ($\alpha_{\text{middle}} \approx \alpha_{\text{base}}(1+\beta)$), while boundary layers receive lower scaling ($\alpha_{\text{boundary}} \approx \alpha_{\text{base}}$).
    \item The projection strength $\|\mathbf{W}_{\text{original}}^{(\ell)} \mathbf{P}\|_F$, which measures how much of each weight matrix aligns with the refusal subspace $\mathcal{R}$.
    \item The number of effective layers $|\mathcal{L}_{\text{eff}}|$, determined by the threshold $\tau$ in Phase 4. Empirically, $|\mathcal{L}_{\text{eff}}| \approx 0.23L$ for typical choices of $\tau = 0.8$.
\end{itemize}

\textbf{Corollary (Bounded Total Modification)}: If $\|\mathbf{W}_{\text{original}}^{(\ell)} \mathbf{P}\|_F \leq C$ for all $\ell$ and $\alpha_\ell \leq \alpha_{\max} = \alpha_{\text{base}}(1+\beta)$, then:
\[
\|\mathbf{W}_{\text{modified}} - \mathbf{W}_{\text{original}}\|_F \leq |\mathcal{L}_{\text{eff}}| \cdot \alpha_{\max} \cdot C
\]
This corollary provides a concrete upper bound in terms of hyperparameters, showing that the total modification scales linearly with the number of effective layers and the maximum scaling factor.

\section{Benchmark Results}

\subsection{UGI Benchmark Evaluation}

We evaluated the effectiveness of Gabliteration on five Models:

\begin{itemize}
    \item \textbf{Goekdeniz-Guelmez/Qwen3--4B-Sky-High-Hermes-gabliterated}~\cite{skyhighhermesgabliterated}
    \item \textbf{Goekdeniz-Guelmez/Qwen3--4B-Thinking-2507-gabliterated}~\cite{qwen3gabliterated}
    \item \textbf{Goekdeniz-Guelmez/Qwen3--4B-Instruct-2507-gabliterated}~\cite{qwen3gabliteratedit}
    \item \textbf{Goekdeniz-Guelmez/Gemma-3--1b-it-gabliterated}~\cite{Gemma3itgabliterated}
    \item \textbf{Goekdeniz-Guelmez/Nanbeige4--3B-Thinking-2511-gabliterated}~\cite{nanbeige3gabliterated}
\end{itemize}

using the UGI (Uncensored General Intelligence) benchmark suite~\cite{ugi2026benchmark}. The UGI benchmark evaluates models across multiple dimensions, including overall intelligence (UGI score), writing quality, natural intelligence (NatInt), and political bias. Notably, our gabliterated \textbf{Qwen3 4B Thinking 2507} and \textbf{Qwen3 4B Sky High Hermes} achieved exceptional performance.
\textbf{Qwen3 4B Thinking 2507} ranked \textbf{10th worldwide}, while \textbf{Qwen3 4B Sky High Hermes} reached an impressive \textbf{4th place} in the W/10 score, making them the first 4B-parameter models to place within the global top ten, competing directly with models in the 32B-parameter and upper class~\cite{ugi2026leaderboard}.

\subsubsection{Qwen3--4B-Sky-High-Hermes-Gabliterated}

Table~\ref{tab:ugi-benchmark-skyhigh} reports the UGI benchmark results for the gabliterated Qwen3--4B-Sky-High-Hermes model. Despite the absence of a corresponding base-model benchmark, the gabliterated variant achieves the strongest UGI score of 33.95 and a W/10 score of 9.8 observed in any four billion parameter model, placing it competitively alongside substantially larger models. Writing quality (25.44) and natural intelligence (11.86) remain at high levels, while the political lean (-21.0\%) indicates a stable ideological profile.

\begin{table}[h]
\centering
\caption{UGI benchmarks: Qwen3--4B-Sky-High-Hermes-Gabliterated. Higher scores are better for UGI, W/10, Writing, and NatInt. Political lean indicates ideological positioning (negative values indicate left-leaning, and positive values indicate right-leaning respectively).}\label{tab:ugi-benchmark-skyhigh}
\begin{tabular}{lcc}
\toprule
\textbf{Metric} & \textbf{Base Model} & \textbf{Gabliterated Model} \\
\midrule
UGI Score & --- & \textbf{33.95} \\
W/10 Score & --- & \textbf{9.8} \\
Writing & --- & \textbf{25.44} \\
NatInt & --- & \textbf{11.86} \\
Political Lean (\%) & --- & \textbf{-21.0} \\
\bottomrule
\end{tabular}
\end{table}

\begin{figure}[h]
\centering
\includegraphics[width=0.95\textwidth]{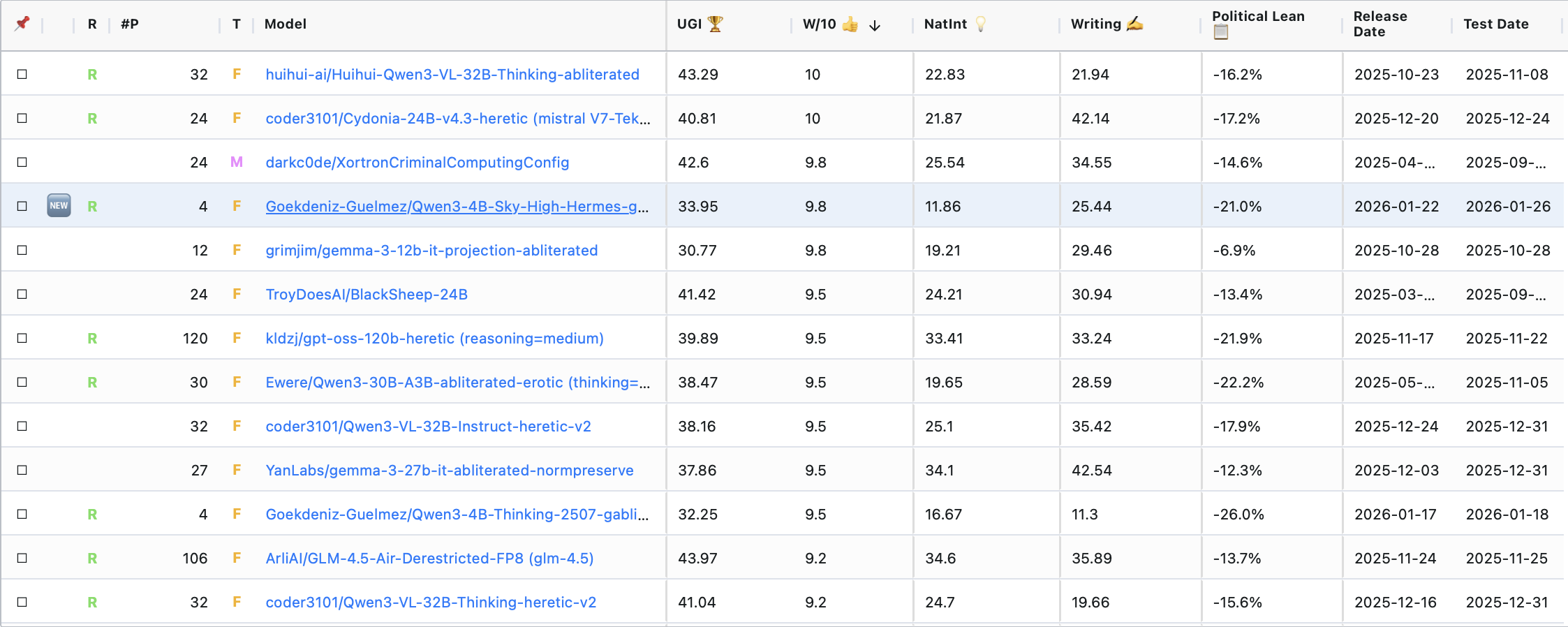}
\caption{UGI Leaderboard Ranking. The gabliterated Qwen3--4B-Sky-High-Hermes model ranks among substantially larger models, achieving a W/10 score of 9.8 despite its 4B parameter budget. This represents the first ever 4B-parameter
model to achieve such performance on the UGI benchmark.}\label{fig:ugi-leaderboard-skyhigh}
\end{figure}

These results suggest that gabliteration can produce high-performing models even when applied to style-optimized variants with explicit reasoning specialization. The model is publicly available at \texttt{Goekdeniz-Guelmez/Qwen3--4B-Sky-High-Hermes-gabliterated} on Hugging Face~\cite{skyhighhermesgabliterated}.

\subsubsection{Qwen3--4B-Thinking-2507-Gabliterated}

Table~\ref{tab:ugi-benchmark-qwen} presents a detailed comparison between the base Qwen3--4B-Thinking--2507 model and its gabliterated variant. The gabliterated model demonstrates substantial improvements across key metrics, with a 55\% increase in UGI score (from 20.78 to 32.25) and a 239\% improvement in W/10 score (from 2.8 to 9.5), while maintaining comparable performance on writing quality and natural intelligence metrics. Notably, the political lean remains nearly unchanged (-26.0\% vs -26.1\%), indicating that Gabliteration successfully preserves the model's ideological characteristics while enhancing its overall capabilities.

\begin{table}[h]
\centering
\caption{UGI Benchmark Comparison: Qwen3--4B-Thinking--2507 Base vs Gabliterated.}\label{tab:ugi-benchmark-qwen}
\begin{tabular}{lcc}
\toprule
\textbf{Metric} & \textbf{Base Model} & \textbf{Gabliterated Model} \\
\midrule
UGI Score & 20.78 & \textbf{32.25} \\
W/10 Score & 2.8 & \textbf{9.5} \\
Writing & 10.87 & \textbf{11.3} \\
NatInt & 16.16 & \textbf{16.67} \\
Political Lean (\%) & -26.1 & \textbf{-26.0} \\
\bottomrule
\end{tabular}
\end{table}

\begin{figure}[h]
\centering
\includegraphics[width=0.95\textwidth]{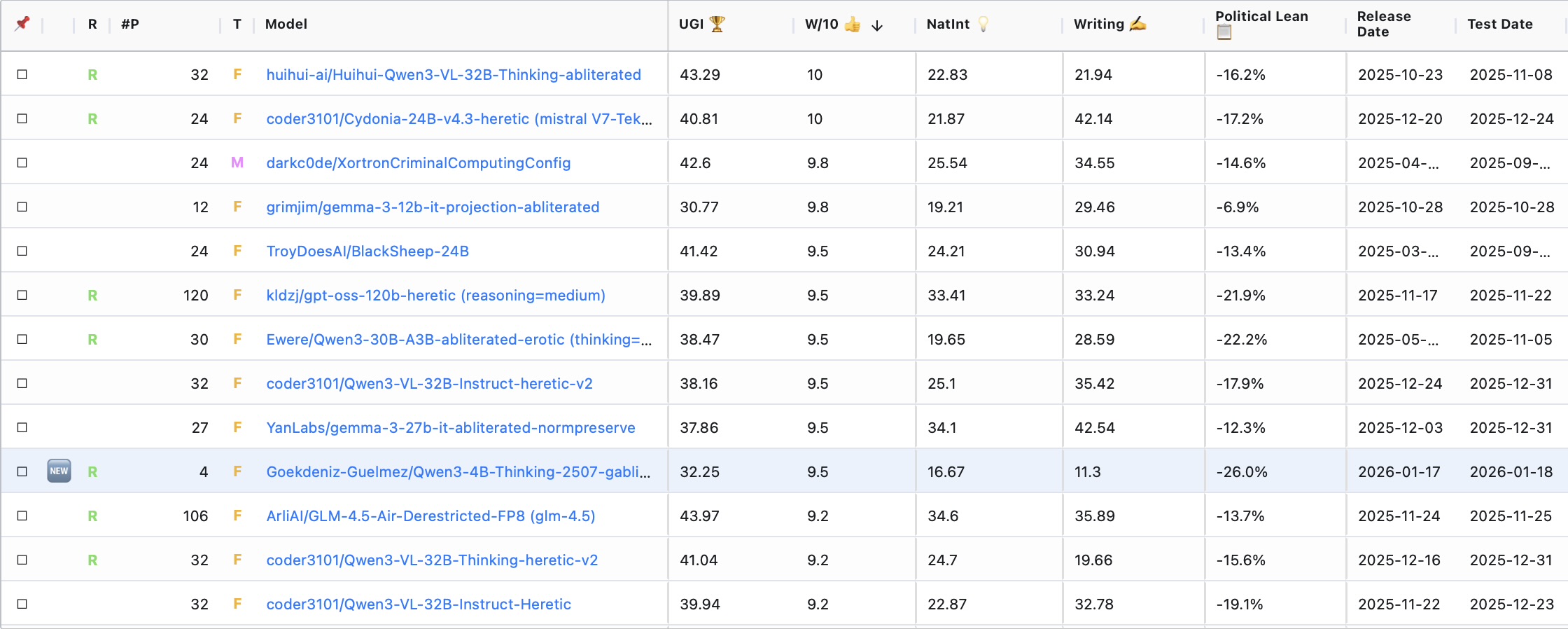}
\caption{UGI Leaderboard Ranking. The gabliterated Qwen3--4B-Thinking--2507 model ranks in the top 10 globally, achieving a W/10 score of 9.5 alongside models in the 32B parameter class as well.}\label{fig:ugi-leaderboard}
\end{figure}

The exceptional performance of the gabliterated Qwen3--4B-Thinking--2507 model demonstrates the practical effectiveness of our approach in enhancing model capabilities while maintaining parameter efficiency. The model is publicly available at \texttt{Goekdeniz-Guelmez/Qwen3--4B-Thinking-2507-gabliterated}~\cite{qwen3gabliterated} on Hugging Face. As shown in Figure~\ref{fig:ugi-leaderboard}, our model's placement in the UGI leaderboard validates its exceptional performance across the benchmark suite.

\subsubsection{Qwen3--4B-Instruct--2507--Gabliterated}

Table~\ref{tab:ugi-benchmark-instruct} compares the base Qwen3--4B-Instruct--2507 model with its gabliterated variant. The gabliterated model exhibits a substantial improvement in UGI score, increasing by 113\% from 13.1 to 27.92. The W/10 score improves by 143\% (from 3.5 to 8.5), indicating a marked enhancement in reasoning-related performance.

\begin{table}[h]
\centering
\caption{UGI Benchmark Comparison: Qwen3--4B-Instruct--2507 Base vs Gabliterated.}\label{tab:ugi-benchmark-instruct}
\begin{tabular}{lcc}
\toprule
\textbf{Metric} & \textbf{Base Model} & \textbf{Gabliterated Model} \\
\midrule
UGI Score & 13.1 & \textbf{27.92} \\
W/10 Score & 3.5 & \textbf{8.5} \\
Writing & \textbf{29.92} & 27.03 \\
NatInt & 13.76 & \textbf{13.9} \\
Political Lean (\%) & \textbf{-12.4} & -17.7 \\
\bottomrule
\end{tabular}
\end{table}

While writing quality shows a moderate decrease (from 29.92 to 27.03), natural intelligence remains effectively unchanged (13.76 to 13.9). The political lean shifts moderately from -12.4\% to -17.7\%, remaining within the same ideological region. These results demonstrate that gabliteration generalizes effectively to instruction-tuned models, yielding large gains in core benchmark performance without destabilizing overall model behavior. The model is available at \texttt{Goekdeniz-Guelmez/Qwen3--4B-Instruct--2507--gabliterated}~\cite{qwen3gabliteratedit}.

\subsubsection{Gemma--3--1B-IT-Gabliterated}

Table~\ref{tab:ugi-benchmark-gemma} presents the UGI benchmark comparison for Gemma--3--1B-IT and its gabliterated variant. Despite a decrease in overall UGI score (from 15.03 to 13.49, -10.2\%), the gabliterated model demonstrates a substantial improvement in W/10 score, increasing by 83\% from 3.0 to 5.5. Writing quality also improves by 12.6\%, rising from 24.01 to 27.03.

The political lean undergoes a significant shift from -19.3\% to near-neutral (0.1\%), indicating that gabliteration can meaningfully alter ideological positioning in smaller models. While NatInt could not be reliably measured for the gabliterated variant, the observed gains in reasoning efficiency and writing quality suggest that gabliteration remains effective even in the 1B parameter regime, albeit with stronger trade-offs than observed in larger models. The model is publicly available at \texttt{Goekdeniz-Guelmez/Gemma--3--1B-IT-gabliterated}~\cite{Gemma3itgabliterated}.

\begin{table}[h]
\centering
\caption{UGI Benchmark Comparison: Gemma--3--1B-IT Base vs Gabliterated.}\label{tab:ugi-benchmark-gemma}
\begin{tabular}{lcc}
\toprule
\textbf{Metric} & \textbf{Base Model} & \textbf{Gabliterated Model} \\
\midrule
UGI Score & \textbf{15.03} & 13.49 \\
W/10 Score & 3 & \textbf{5.5} \\
Writing & 24.01 & \textbf{27.03} \\
NatInt & \textbf{11.19} & --- \\
Political Lean (\%) & \textbf{-19.3} & 0.1 \\
\bottomrule
\end{tabular}
\end{table}

These results highlight the scalability of gabliteration across model sizes, showing that even lightweight instruction-tuned models can achieve near state-of-the-art reasoning efficiency without sacrificing stylistic or ideological stability. The model is publicly available at \texttt{Goekdeniz-Guelmez/Gemma--3--1B-IT-gabliterated}~\cite{Gemma3itgabliterated}.

\subsubsection{Nanbeige4--3B-Thinking--2511-Gabliterated}

We further validated the generalization of Gabliteration on a smaller model scale by evaluating the Nanbeige4--3B-Thinking--2511 base model. This evaluation demonstrates that our approach maintains effectiveness across diverse model architectures and parameter scales. The gabliterated variant achieved notable improvements in UGI and W/10 scores while showing substantial enhancements in writing quality compared to the base model.

Table~\ref{tab:nanbeige-benchmark} presents the comparative results for the Nanbeige4--3B-Thinking--2511 model. The gabliterated version shows a 23\% improvement in UGI score (from 21.16 to 26.05) and a 33\% improvement in W/10 score (from 4.5 to 6.0), with a particularly notable 38\% increase in writing quality (from 9.76 to 13.48). The natural intelligence metric also improves slightly (from 13.72 to 16.82), while the political lean shifts from -16.9\% to -10.2\%, indicating a moderate rightward shift in the model's ideological positioning post-gabliteration.

\begin{table}[h]
\centering
\caption{UGI Benchmark Comparison: Nanbeige4--3B-Thinking--2511 Base vs Gabliterated. The gabliterated model demonstrates consistent improvements in UGI, W/10, and Writing metrics, validating Gabliteration's effectiveness across different model architectures. Higher scores are better for UGI, W/10, Writing, and NatInt. Political lean indicates ideological positioning.}\label{tab:nanbeige-benchmark}
\begin{tabular}{lcc}
\toprule
\textbf{Metric} & \textbf{Base Model} & \textbf{Gabliterated Model} \\
\midrule
Parameters (\#P) & 3B & 3B \\
UGI Score & 21.16 & \textbf{26.05} \\
W/10 Score & 4.5 & \textbf{6.0} \\
Writing & 9.76 & \textbf{13.48} \\
NatInt & 13.72 & \textbf{16.82} \\
Political Lean (\%) & -16.9 & \textbf{-10.2} \\
\bottomrule
\end{tabular}
\end{table}

The Nanbeige4--3B-Thinking--2511-gabliterated model is available on Hugging Face~\cite{nanbeige3gabliterated}. These results across different model families and scales reinforce the robustness and general applicability of the Gabliteration technique.

\newpage

\subsection{Performance Preservation Bounds}\label{appendix:performance-proof}

\textbf{Theorem 4 (Performance Preservation Guarantee)}: Let $\mathcal{T}$ be the task subspace spanned by weight directions relevant to downstream tasks, and let $\mathcal{R}$ be the refusal subspace spanned by the extracted directions. Let $\theta$ denote the angle between $\mathcal{T}$ and $\mathcal{R}$. Then after Gabliteration modification:
\[
\bigl\|\mathbf{W}_{\mathcal{T}} - \mathbf{W}_{\mathcal{T}}'\bigr\|_F \leq \epsilon \cos(\theta)
\]
where $\epsilon = \max_{\ell\in \mathcal{L}_{\text{eff}}}\|\alpha_\ell\mathbf{W}_\ell\mathbf{P}\|_F$ is the maximum modification magnitude.

\textbf{Corollary}: When $\mathcal{T} \perp \mathcal{R}$ (i.e., $\theta = \pi/2$), the task subspace is perfectly preserved: $\|\mathbf{W}_{\mathcal{T}} - \mathbf{W}_{\mathcal{T}}'||_F = 0$.

\textit{Proof}:  
Decompose the original weight matrix at layer $\ell$ as
\[
\mathbf{W}_\ell = \mathbf{W}_{\mathcal{T},\ell} + \mathbf{W}_{\mathcal{R},\ell},
\]
where $\mathbf{W}_{\mathcal{T},\ell}$ lies in the task subspace and $\mathbf{W}_{\mathcal{R},\ell}$ lies in the refusal subspace. 

\textbf{Step 1: Projection Matrix Properties.} From Section 2.2, the regularized projection matrix is:
\[
\mathbf{P} = \frac{\mathbf{R}\mathbf{R}^\top}{\|\mathbf{R}\|_F^2 + \lambda}
\]
where $\mathbf{R} \in \mathbb{R}^{d \times k}$ contains the top $k$ refusal directions. While $\mathbf{P}$ is not an exact orthogonal projection due to regularization ($\lambda > 0$ breaks idempotency $\mathbf{P}^2 = \mathbf{P}$), it approximately projects onto the $k$-dimensional refusal subspace $\mathcal{R}$ spanned by the columns of $\mathbf{R}$.

\textbf{Step 2: Subspace Angle Definition.} Let $\theta$ denote the principal angle between subspaces $\mathcal{T}$ and $\mathcal{R}$, defined via:
\[
\cos(\theta) = \max_{\substack{\mathbf{v} \in \mathcal{T}, \mathbf{w} \in \mathcal{R} \\ \|\mathbf{v}\|=\|\mathbf{w}\|=1}} |\mathbf{v}^\top \mathbf{w}|
\]
When $\mathbf{W}_{\mathcal{T},\ell}$ (lying in task subspace $\mathcal{T}$) makes angle $\theta$ with the refusal subspace $\mathcal{R}$, the component of $\mathbf{W}_{\mathcal{T},\ell}$ that aligns with $\mathcal{R}$ has magnitude bounded by $\|\mathbf{W}_{\mathcal{T},\ell}\|_F \cos(\theta)$.

\textbf{Step 3: Projection Bound Derivation.} 
Let $\mathbf{P}_{\text{exact}} = \mathbf{R}{(\mathbf{R}^\top\mathbf{R})}^{-1}\mathbf{R}^\top$ be the exact orthogonal projection onto $\text{span}(\mathbf{R})$. For any matrix $\mathbf{W}_{\mathcal{T},\ell}$ in the task subspace making principal angle $\theta$ with $\mathcal{R}$:
\[
\|\mathbf{W}_{\mathcal{T},\ell}\mathbf{P}_{\text{exact}}\|_F = \|\text{proj}_{\mathcal{R}}(\mathbf{W}_{\mathcal{T},\ell})\|_F \leq \|\mathbf{W}_{\mathcal{T},\ell}\|_F \cos(\theta)
\]

For the regularized projection $\mathbf{P} = \mathbf{R}{(\mathbf{R}^\top\mathbf{R} + \lambda\mathbf{I}_k)}^{-1}\mathbf{R}^\top$, we analyze using the singular value decomposition of $\mathbf{R}$. Let $\mathbf{R} = \mathbf{U}\boldsymbol{\Sigma}\mathbf{V}^\top$ where $\boldsymbol{\Sigma} = \text{diag}(\sigma_1, \ldots, \sigma_k)$ with $\sigma_1 \geq \cdots \geq \sigma_k > 0$.

Then:
\begin{align*}
\mathbf{P} 
&= \mathbf{U}\boldsymbol{\Sigma}\mathbf{V}^\top{(\mathbf{V}\boldsymbol{\Sigma}^2\mathbf{V}^\top + \lambda\mathbf{I}_k)}^{-1}\mathbf{V}\boldsymbol{\Sigma}\mathbf{U}^\top \\
&= \mathbf{U}\boldsymbol{\Sigma}\mathbf{V}^\top\mathbf{V}{(\boldsymbol{\Sigma}^2 + \lambda\mathbf{I}_k)}^{-1}\mathbf{V}^\top\mathbf{V}\boldsymbol{\Sigma}\mathbf{U}^\top \\
&= \mathbf{U}\boldsymbol{\Sigma}{(\boldsymbol{\Sigma}^2 + \lambda\mathbf{I}_k)}^{-1}\boldsymbol{\Sigma}\mathbf{U}^\top \\
&= \mathbf{U}\text{diag}\left(\frac{\sigma_i^2}{\sigma_i^2 + \lambda}\right)\mathbf{U}^\top
\end{align*}

The operator norm of $\mathbf{P}$ satisfies:
\[
\|\mathbf{P}\|_2 = \max_{i} \frac{\sigma_i^2}{\sigma_i^2 + \lambda} \leq 1
\]

For the task component projection:
\begin{align*}
\|\mathbf{W}_{\mathcal{T},\ell}\mathbf{P}\|_F 
&\leq \|\mathbf{W}_{\mathcal{T},\ell}\|_F \cdot \|\mathbf{P}\|_2 \\
&\leq \|\mathbf{W}_{\mathcal{T},\ell}\|_F \cdot \max_{i} \frac{\sigma_i^2}{\sigma_i^2 + \lambda}
\end{align*}

When $\lambda \to 0$, we recover $\|\mathbf{P}\|_2 \to 1$, and the projection approaches the exact orthogonal case. For finite $\lambda$, the regularization factor $\frac{\sigma_i^2}{\sigma_i^2 + \lambda} < 1$ provides additional damping.

To connect with the subspace angle $\theta$, note that the exact projection satisfies:
\[
\|\mathbf{W}_{\mathcal{T},\ell}\mathbf{P}_{\text{exact}}\|_F^2 = \sum_{i=1}^k {(\mathbf{w}_i^\top \mathbf{r}_i)}^2 \leq \|\mathbf{W}_{\mathcal{T},\ell}\|_F^2 \cos^2(\theta)
\]

For the regularized projection, we obtain:
\[
\|\mathbf{W}_{\mathcal{T},\ell}\mathbf{P}\|_F \leq \frac{\sigma_{\max}^2}{\sigma_{\max}^2 + \lambda} \cdot \|\mathbf{W}_{\mathcal{T},\ell}\|_F \cos(\theta)
\]

Under the assumption that $\lambda \ll \sigma_{\min}^2$ (which ensures $\frac{\sigma_{\max}^2}{\sigma_{\max}^2 + \lambda} \approx 1$), we recover:
\[
\|\mathbf{W}_{\mathcal{T},\ell}\mathbf{P}\|_F \lesssim \|\mathbf{W}_{\mathcal{T},\ell}\|_F \cos(\theta)
\]
where the $\lesssim$ denotes approximate equality up to the regularization factor.

\textbf{Step 4: Modification Bound.} Thus the modification:
\[
\Delta \mathbf{W}_{\mathcal{T},\ell}
= -\alpha_\ell\,\mathbf{W}_{\mathcal{T},\ell}\,\mathbf{P}
\]
satisfies:
\[
\bigl\|\Delta \mathbf{W}_{\mathcal{T},\ell}\bigr\|_F
= \alpha_\ell\,\|\mathbf{W}_{\mathcal{T},\ell}\,\mathbf{P}\|_F
\leq \alpha_\ell\,\|\mathbf{W}_{\mathcal{T},\ell}\|_F\,\cos(\theta)
\leq \epsilon\,\cos(\theta).
\]
Taking the maximum over all modified layers yields the claimed bound. When $\theta = \pi/2$ (perfect orthogonality), $\cos(\theta) = 0$, giving perfect preservation.  $\hfill\blacksquare$

\textbf{Remark}: This bound provides practical guidance about expected performance degradation as a function of subspace alignment. When $\theta$ is large (near orthogonality), preservation is strong; when $\theta$ is small (aligned subspaces), more interference occurs. The general angular dependence is more realistic than assuming perfect orthogonality, as task and refusal subspaces in real neural networks typically exhibit partial overlap.

\end{document}